\documentclass[twoside]{article}

\usepackage[accepted]{aistats2024}

\usepackage[utf8]{inputenc} 
\usepackage[T1]{fontenc}    
\usepackage{url}            
\usepackage{booktabs}       
\usepackage{amsfonts}       
\usepackage{nicefrac}       
\usepackage{microtype}      
\usepackage{xcolor}         
\usepackage{amsmath}
\usepackage{algorithm2e}
\usepackage[hidelinks]{hyperref} 
\usepackage{cleveref}
\usepackage{amsthm}
\usepackage{amsmath}
\usepackage{cancel}
\usepackage{mwe}
\usepackage{graphicx}
\usepackage{ragged2e}
\usepackage{subcaption}
\usepackage{todonotes}
\usepackage{makecell}
\usepackage{setspace}
\usepackage{mwe}
\usepackage[citestyle=authoryear,natbib=true,backend=bibtex,doi=false,isbn=false,url=false,eprint=false]{biblatex}

\newtheorem{proposition}{Proposition}

\newtheorem*{proposition_star}{Proposition}

\newtheorem{innercustomgeneric}{\customgenericname}
\providecommand{\customgenericname}{}
\newcommand{\newcustomtheorem}[2]{%
  \newenvironment{#1}[1]
  {%
   \ifdefined\crefalias\crefalias{innercustomgeneric}{#2}\fi
   \renewcommand\customgenericname{#2}%
   \renewcommand\theinnercustomgeneric{##1}%
   \innercustomgeneric
  }
  {\endinnercustomgeneric}%
  \ifdefined\crefname\crefname{#2}{#2}{#2s}\fi
}

\newcustomtheorem{customthm}{Theorem}
\newcustomtheorem{customlemma}{Lemma}
\newcustomtheorem{customprop}{Proposition}

\SetCommentSty{mycommfont}

\begin{document}

%

%

\twocolumn[

\aistatstitle{Sequential Monte Carlo for Inclusive KL Minimization in Amortized Variational Inference}

\aistatsauthor{Declan McNamara \And Jackson Loper \And  Jeffrey Regier}

\aistatsaddress{University of Michigan \And  University of Michigan \And University of Michigan} ]
\begin{abstract}
  For training an encoder network to perform amortized variational inference, the Kullback-Leibler (KL) divergence from the exact posterior to its approximation, known as the inclusive or forward KL, is an increasingly popular choice of variational objective due to the mass-covering property of its minimizer. However, minimizing this objective is challenging. A popular existing approach, Reweighted Wake-Sleep (RWS), suffers from heavily biased gradients and a circular pathology that results in highly concentrated variational distributions. As an alternative, we propose SMC-Wake, a procedure for fitting an amortized variational approximation that uses likelihood-tempered sequential Monte Carlo samplers to estimate the gradient of the inclusive KL divergence. We propose three gradient estimators, all of which are asymptotically unbiased in the number of iterations and two of which are strongly consistent. Our method interleaves stochastic gradient updates, SMC samplers, and iterative improvement to an estimate of the normalizing constant to reduce bias from self-normalization. In experiments with both simulated and real datasets, SMC-Wake fits variational distributions that approximate the posterior more accurately than existing methods.
\end{abstract}

\section{INTRODUCTION}
\label{intro}

Amortized variational inference (VI) fits an encoder $q_\phi(z \mid x)$ to approximate the posterior for each $x$ in a dataset $\mathcal{D} = \{x_1,\dots, x_n\}$. The encoder parameters $\phi$ are selected to minimize some divergence between the variational posteriors $q_\phi(z \mid x)$ and the exact posteriors $p(z \mid x)$, averaged over observations $x \in \mathcal{D}$. The forward KL divergence (\autoref{forward-kl}), also known as the inclusive KL divergence, is an increasingly popular variational objective because minimizers $q_\phi$ tend to be mass covering (overdispersed) with respect to the true posterior \citep{Gu2015NeuralAdaptiveSMC}; this property is desirable for applications benefiting from conservative uncertainty quantification.  In contrast, reverse KL minimizers tend to be mode-seeking and to underestimate uncertainty \citep{Domke2018IWVI}. 

Reweighted Wake-Sleep (RWS) is a popular method for performing amortized variational inference that attempts to minimize the forward KL divergence, but, surprisingly, in practice wake-phase training can result in variational posteriors that are severely underdispersed \citep{Le2019RevisitRWS}. We conjecture that this behavior results from a ``circular pathology'' that arises from fitting $q_\phi$ with particles proposed from $q_\phi$ itself (\Cref{section:mass_concentration}).

To address these limitations of RWS, we propose SMC-Wake, a method to fit an amortized encoder by minimizing the average forward KL divergence via stochastic gradient descent (SGD) with gradients estimated using sequential Monte Carlo (SMC) samplers (\Cref{background}). We propose three new estimators of the gradient of the forward KL 
 that combine SMC samplers in different ways (\Cref{method}), and we prove that each is asymptotically unbiased and that two are also strongly consistent  (\Cref{subsection:analysis}). Our method compares favorably to those in related work (\Cref{related_work}), and we demonstrate these advantages empirically in a variety of experimental settings (\Cref{experiments}). Compared to other forward KL minimization methods based on importance sampling, SMC-Wake leverages higher-fidelity particle approximations with lower-variance weights, resulting in a stable training regime that avoids weight degeneracy and the circular pathology we identify in RWS. 

\section{BACKGROUND}
\label{background}

\subsection{Reweighted Wake-Sleep}
\label{background:rws}

Given a dataset $\mathcal{D} = \{x_1, \dots, x_n\}$, the wake-phase update to the variational parameters in reweighted wake-sleep (RWS)  \citep{BornscheinBengio2015RWS} aims to minimize the average forward KL divergence:
\begin{equation}
  \label{forward-kl}
\frac{1}{n} \sum_{j=1}^n \textrm{KL}(p(z \mid x_j) \mid \mid q_\phi(z \mid x_j)).
\end{equation}
The gradient of this objective, however, is generally intractable; even a Monte Carlo estimate of it requires sampling from the exact posteriors $p(z \mid x_j)$  for each $x_j \in \mathcal{D}$. Instead, RWS approximates the gradient of each term of \autoref{forward-kl} using self-normalized importance sampling, with the current iterate of $q_\phi(z \mid x)$ serving as the proposal. With $K$ particles, the resulting gradient estimator is
\begin{equation}
  \label{wake_gradient}
  -\sum_{i=1}^K w^i \nabla_\phi \log q_\phi(z^i \mid x), \textrm{ 
 where  } w^i = \frac{\tilde{w}^i}{\sum_j \tilde{w}^j}, 
\end{equation}
$\tilde{w}^i = \frac{p(z^i, x)}{q_\phi(z^i \mid x)}$ are unnormalized importance weights, and $z^1,\dots,z^K \overset{iid}{\sim} q_\phi(z \mid x)$ for a given $x$. Although biased, this estimator converges almost surely to the gradient of the forward KL divergence as $K \to \infty$, under mild conditions \citep{Owen2013ImportanceSampling}. RWS also includes a sleep phase that averages exact gradients of simulated (``dreamt'') data in a likelihood-free approach to inference. However, this approach has its own limitations and has been shown to underperform the wake-phase update in some cases \citep{Le2019RevisitRWS}.

\subsection{Sequential Monte Carlo samplers}
\label{background:particle_filter_smc}

Sequential Monte Carlo (SMC) samplers \citep{DelMoral2006SequentialMonteCarlo} compute estimates of $\mu = \mathbb{E}_{\gamma(z)} f(z)$ for an integrable test function $f$ and a target distribution $\gamma$. SMC samplers generalize self-normalized importance sampling (SNIS) \citep{Owen2013ImportanceSampling}: instead of a single target $\gamma,$ a \textit{sequence} of targets $\gamma_1, \dots, \gamma_T$ is approximated. Even if only a single target $\gamma$ is of interest, SMC samplers can improve estimation through annealing. One selects a sequence of distributions such that $\gamma_1$ is tractable and $\gamma_T = \gamma$, with intermediate distributions that facilitate moves between these \citep{Chopin2020IntroSMC}. 

In a Bayesian setting, where approximations of the posterior are of interest, typically the final target is $\gamma \overset{d}{=} p(z \mid x)$ for some observation $x$. A particular sequence of targets is given by likelihood-tempered sequential Monte Carlo (LT-SMC) \citep{Chopin2020IntroSMC}. LT-SMC runs SMC samplers using a base distribution, typically the prior $p(z)$, as the initial target $\gamma_1$, and then anneals toward $p(z \mid x)$ through the intermediate targets $\gamma_t(z) \propto p(z)p(x \mid z)^{\tau_t}$ with $0 = \tau_1 < \cdots < \tau_T = 1$. 


SMC samplers form \textit{discrete} or \textit{empirical} approximations to each distribution $\gamma_t(z)$. The discrete approximation to $\gamma_1$ is computed by importance sampling, which produces an initial $K$-particle approximation $\textrm{Cat}(z_{1}^{1:K}, w_1^{1:K})$ to $\gamma_1$. Then, at each stage $t=1,\dots, T-1$, three distinct steps are performed:
\begin{equation*}
\begin{split}
  \label{eqn:resample_propagate_update}
  &\textrm{Resample: Draw  } z^{(i)}_{t} \overset{iid}{\sim} \textrm{Cat}(z_{t}^{1:K}, w_t^{1:K}),  \ i \in [K]. \\
  &\textrm{Mutate: Propose  } z^i_{t+1} \sim M(z^{(i)}_{t}, dz_{t+1}), \ i \in [K]. \\
  &\textrm{Update: Recalculate weights  } w_{t+1}^{1:K} \textrm{  for } z_{t+1}^{1:K}.
\end{split}
\end{equation*}

Above, $\textrm{Cat}(\cdot, \cdot)$ denotes a categorical distribution and $M(\cdot, dz)$ denotes a transition kernel. The weight update depends on the transition kernel. Expectations with respect to the posterior $\gamma_T = p(z \mid x)$ can be approximated by Monte Carlo integration, that is, $\mathbb{E}_{\gamma_T} f(z) \approx \sum_{i=1}^K w_T^i f(z_T^i)$. These approximations are biased but consistent as $K \to \infty$ with bias and variance of order $O(\frac{1}{K})$ \citep{Chopin2020IntroSMC}. 

The random variables $z_t^{1:K}$ and $w_t^{1:K}$ for $t < T$ are auxiliary: they are not used to approximate $\mathbb{E}_{p(z \mid x)} f(z)$, but to guide particles toward areas of high mass in the target distribution. Resampling eliminates particles with low weights, while mutation and reweighting provide a means to produce new particles and to weight them according to their quality with respect to the subsequent target $\gamma_{t+1}(z)$.

Like SNIS, SMC only requires the evaluation of an unnormalized density $\Tilde{\gamma}_t(z)$ of each target distribution, as is typically required in a Bayesian setting. Compared to Markov chain Monte Carlo (MCMC) methods, SMC typically requires many fewer steps (e.g., $T < 100$) and is more readily adaptable: for example, the mutation kernel need not satisfy detailed balance conditions \citep{Naesseth2019Elements}. Although SMC is more expensive than SNIS due to the $T > 1$ stages, parallelization of operations across the $K$ particles can be highly efficient.

\section{MASS CONCENTRATION IN RWS}
\label{section:mass_concentration}

To fit an amortized encoder network to minimize the forward KL divergence through iterative optimization, we repeatedly evaluate its gradient:
\begin{equation}
    \label{eqn:forward_kl_gradient}
    \frac{1}{n}\sum_{j=1}^n \mathbb{E}_{p(z \mid x_j)} \nabla_\phi \log q_\phi(z \mid x_j).
\end{equation}
Doing so requires us to evaluate expectations with respect to the posteriors $p(z \mid x_j)$. Estimating each term of \autoref{eqn:forward_kl_gradient} using self-normalized importance sampling (SNIS) with $q_\phi$ as the proposal, as in wake-phase training, can result in degenerate variational distributions that concentrate mass. \citet{Le2019RevisitRWS}  first noted this degeneracy for small $K$, but our case studies show that it persists in practice even for large numbers of particles, such as $K=1000$, and, surprisingly, it is not resolved by taking a defensive approach to importance sampling  (\Cref{experiments}). We conjecture that this failure mode is due to a ``circular pathology'' that arises when simultaneously proposing from $q_\phi$ to optimize the score of $q_\phi$ itself. Consider the wake-phase gradient
\begin{equation*}
    -\sum_{i=1}^K w^i \nabla_\phi \log q_\phi(z^i \mid x)
\end{equation*}
for fixed $x$, with $w^i$ denoting the normalized weights. This gradient can be viewed as a Monte Carlo estimate of the gradient of the surrogate objective
\begin{equation}
    \label{eqn:surrogate_objective}
    \mathbb{E}_{z^1,\dots,z^K \overset{iid}{\sim} q_{\phi}(z \mid x)} -\sum_{i=1}^K w^i \log q_\phi(z^i \mid x)
\end{equation}
 if one places stop gradient operators on the weights and the law used to generate samples (refer to \Cref{appendix:circular} for more details). We now investigate the properties of this objective. Note that wake-phase updates do not directly optimize \autoref{eqn:surrogate_objective}, but the wake-phase gradient estimator is computed identically to a Monte Carlo estimator of the gradient of \autoref{eqn:surrogate_objective} with stop gradient operations in place. It is illustrative to consider what might go wrong if one attempted to minimize this surrogate objective directly, as wake-phase training dynamics may behave similarly.

Let $\mathcal{L}(q_\phi)$ denote the value of the surrogate functional above for any given proposal distribution $q_\phi$. When $K=1$ we note that this objective can made arbitrarily low by choosing a value of $q$ that is quite peaked (and also arbitrarily different from the true posterior).  For any $x$, a sample $z' \sim q_\phi(z \mid x)$ is drawn and the objective is approximated using the log of its own density, i.e. $\log q_\phi(z' \mid x)$. Thus, a highly peaked $q_\phi$ can trivially achieve an arbitrarily low value of the objective. This result should not be so surprising: SNIS fails when applied with only a single sample.  As is well known, SNIS also fails when the target is not absolutely continuous with respect to the proposal.  As a result, even taking $K$ to be arbitrarily large, it is possible to find peaked proposals that are arbitrarily dissimilar from the true posterior and yet lead to arbitrarily low values of the surrogate objective.

\begin{proposition}
\label{prop:formal_circular}
    Let $\mathcal{L}(q)$ denote the surrogate objective defined above for fixed $x$ and fixed $K \in \mathbb{N}$. Let $p$ denote the posterior $p(z \mid x)$. Then there exists $q(z) \neq p(z \mid x)$ such that $\mathcal{L}(q) < \mathcal{L}(p)$.
\end{proposition}

\Cref{appendix:circular} provides a proof. \Cref{prop:formal_circular} implies that even for arbitrarily large $K$, there exists a $q$ with a lower surrogate objective value than the exact posterior has. Our proof constructs $q \sim \mathrm{Unif}(0,\delta)$ for small $\delta$.

In practice, the same issue arises even when $p(z \mid x)$ is absolutely continuous with respect to $q_\phi$, as demonstrated by the following toy example.  Take $z \sim \mathcal{N}(0, 10^2)$ and $x \mid z \sim \mathcal{N}(z, 1^2)$. The exact posterior is $\mathcal{N}(\frac{100}{101}x, \frac{100}{101})$. Using $K=10,000$ importance samples, we estimate the surrogate objective for each of the highly peaked Gaussian variational distributions in \Cref{tab:gaussian_toy_circular_pathology} for a fixed, simulated draw $x$. The estimated objective for each peaked Gaussian is lower than the estimated objective for the exact posterior.

\begin{table}[ht!]
    \centering
    \begin{tabular}{lr}
\toprule
Proposal Distribution & Wake Objective \\
\midrule
$q \sim \mathcal{N}(0,.0001^2)$ & -4.690 (1.471) \\
$q \sim \mathcal{N}(0,.00001^2)$ & -6.841 (1.947) \\
$q \sim \mathcal{N}(0,.000001^2)$ & -9.439 (1.497) \\
$q \sim \mathcal{N}(0,.0000001^2)$ & -11.798 (1.585) \\
\hline
$q \sim \mathcal{N}(\frac{100}{101}x, \frac{100}{101})$ & 1.415 (0.008) \\
\bottomrule
\end{tabular}
    \caption{Avg. surrogate objective values for highly-peaked Gaussian proposals (standard errors in parentheses).}
    \label{tab:gaussian_toy_circular_pathology}
\end{table}
\raggedbottom

Using $q_\phi$ as a proposal for SNIS (as in RWS) is common, perhaps because there is a known case in which doing so results in unbiased SNIS estimates: if $q_\phi(z \mid x_j) = p(z \mid x_j)$ for all $x_j \in \mathcal{D}$, the SNIS estimate of \autoref{eqn:forward_kl_gradient} is unbiased. However, this special case is likely to be irrelevant in real-world settings: firstly, there may be no parameters for which this equality holds due to the variational gap and the ``amortization gap'' \citep{Cremer2018AmortizationGap}, and secondly, the circular pathology itself may prevent optimization trajectories from converging to such parameter values. The circular pathology can be remedied by using an SNIS proposal that does not depend on $q_\phi$, such as the prior $p(z)$. However, the prior is largely uninformative, so most of the particles sampled from it will be of poor quality. 

\section{SMC-WAKE} 
\label{method}

To address the deficiencies of RWS, we propose a method called SMC-Wake for fitting $q_\phi(z \mid x)$ to minimize the average forward KL divergence (\autoref{forward-kl}). Our method is detailed in \Cref{algorithm:smc_wake}. SMC-Wake generates discrete particle approximations $\hat{P}$ to the posterior $p(z \mid x_j)$ for each point $x_j \in \mathcal{D}$, and fits the encoder $q_\phi(z \mid x)$ by using these to estimate the gradient of the forward KL divergence. The proposed method uses sequential Monte Carlo samplers to construct discrete particle approximations to the posterior, rather than importance samplers. SMC-Wake also uses a gradient estimator that is consistent as the number of SMC sampler runs tends to infinity (with a fixed number of SMC particles per sampler).

\subsection{Our LT-SMC subroutine}
\label{subsection:resolving_circular_pathology}
For each point $x_j \in \mathcal{D}$, SMC-Wake constructs discrete particle approximations $\hat{P}$ to $p(z \mid x_j)$ using LT-SMC (sketched in \Cref{algorithm:LT-SMC_short}, \Cref{appendix:algo}). We perform mutation at each stage using a random-walk Metropolis-Hastings kernel $M_t(\cdot, dz_t)$ that is invariant with respect to $\gamma_{t-1}$. Our exact implementation (\Cref{appendix:algo}) follows that of \citet{Chopin2020IntroSMC} and uses adaptive temperature selection and optional resampling. Given an observation $x_j$ and a number of particles $K$, LT-SMC provides a discrete approximation $\textrm{Cat}(z_T^{1:K}, w_T^{1:K})$ to the posterior $p(z \mid x_j)$, which we use to estimate a term of the gradient of the forward KL divergence (\autoref{eqn:forward_kl_gradient}). In our notation, we henceforth suppress the dependence on $T$, as we always use discrete particle approximations from the final stage $T$ to compute expectations. Importantly, the entire LT-SMC procedure does not depend on $q_\phi$. By using the prior as the base distribution $\gamma_1$, LT-SMC avoids the circular pathology that we previously identified (\Cref{section:mass_concentration}). Although the prior alone is uninformative as a proposal, annealing along a tempering schedule results in discrete particle approximations that approximate the per-observation posteriors $p(z \mid x_j)$ well.

\subsection{Consistent gradient estimation}
\label{subsection:gradient_estimators}
For any fixed number of particles $K$ and observation $x_j \in \mathcal{D}$, the particle approximations from LT-SMC yield gradient estimators $-\sum_{i=1}^K w^i\nabla_\phi \log q_\phi(z^i \mid x_j)$, which are biased for $\mu_j = -\mathbb{E}_{p(z \mid x_j)}\nabla_\phi \log q_\phi(z \mid x_j)$. This is problematic because stochastic gradient descent based on biased estimators is not guaranteed to converge.  In this section, we propose three gradient estimators that are asymptotically unbiased, including two that are consistent. Our estimators accomplish this by reducing the bias from self-normalization by averaging the different particle approximations that are computed across the optimizer iterations. Asymptotic results are obtained as the number of particle approximations $M \to \infty$, where the number of particles $K$ remains fixed.

Consider a fixed observation $x$. The final target $\gamma_T(z) \propto p(z,x)$ in LT-SMC is known up to a normalization constant $C$ such that $p(z,x) = C \gamma_T(z)$, i.e., $C = p(x)$. Each individual LT-SMC sampler yields an estimate
\begin{equation}
    \label{eqn:evidence_estimator}
    \hat{C} = \prod_{t=1}^T \bigg{(} \frac{1}{K} \sum_{i=1}^K \Tilde{w}_t^i \bigg{)},
\end{equation}
where $\tilde{w}_t^{1:K}$ are the unnormalized weights from \Cref{algorithm:LT-SMC_short}. This estimate satisfies $\mathbb{E} \hat{C} = C$, with the expectation taken over all the random variables generated by the SMC algorithm \citep{Naesseth2019Elements}. SMC-Wake constructs an estimate of $C$ from $M$ LT-SMC runs using the estimator $\frac{1}{M} \sum_{m=1}^M \hat{C}_{(m)}$, where $\hat{C}_{(m)}$ is the estimate of the normalization constant from iteration $m \in \{1,\ldots, M\}$. By the strong law of large numbers, $\frac{1}{M} \sum_{m=1}^M \hat{C}_{(m)} \overset{a.s.}{\to} C$ as $M \to \infty$.

SMC samplers can provide unbiased estimates of $C \mathbb{E}_{p(z \mid x)} f(z)$ for a test function $f$. As $C = p(x)$ is a high-dimensional integral, $C$ is usually unknown and so one typically divides by the estimate $\hat C$, resulting in bias. Within our iterative fitting procedure for $q_\phi$, we instead propose to combine many normalization constant estimates produced by many runs of SMC. These can be averaged with $O(1)$ memory to iteratively refine our estimate of $C$ while simultaneously fitting the variational parameters $\phi$. As the number of LT-SMC runs $M$ increases, $\frac{1}{M} \sum_{m=1}^M \hat{C}_{(m)}$ converges to $C$, leading to unbiased estimates of $\mathbb{E}_{p(z \mid x)} f(z)$ as $M \to \infty$.

\Cref{tab:estimators} provides three different ratio estimators, constructed with $f(z)= -\nabla_\phi \log q_\phi(z \mid x)$ as the test function for estimating the gradient of the forward KL divergence. In each case, $z_{(m)}^{1:K}$, $w_{(m)}^{1:K}$, and $\hat C_{(m)}$ denote the atom positions, weights, and normalizing constant estimates returned by the $m$th run of LT-SMC.  For $\hat \nabla^{(b)}$, $\tilde z_{(m)}$ refers to a single draw from $\mathrm{Cat}(z_{(m)}^{1:K},w_{(m)}^{1:K})$.  Each of the numerators in \Cref{tab:estimators} is an unbiased estimator of $C \mathbb{E}_{p(z \mid x)} f(z)$. Dividing each by $\frac{1}{M} \sum_{m=1}^M \hat{C}_{(m)}$ yields an estimator of $\mathbb{E}_{p(z \mid x)} f(z)$. The different numerators exploit different aggregations of the empirical distributions $\textrm{Cat}(z_{(m)}^{1:K}, w_{(m)}^{1:K})$ to approximate the posterior $p(z \mid x)$, each with different memory requirements.

\begin{table}[ht!]
\caption{Forward KL Gradient Estimators}
    \label{tab:estimators}
\renewcommand{\arraystretch}{2}
\small
    \centering
    \begin{tabular}{c|c|c}
        Label & Estimator & Memory  \\
        \hline
       $\hat{\nabla}^{(a)}$ & $\frac{\frac{1}{M}\sum_{m=1}^M \hat{C}_{(m)} \big{(} \sum_{k=1}^K w^k_{(m)} f(z^k_{(m)}) \big{)}}{\frac{1}{M} \sum_{m=1}^M \hat{C}_{(m)}}$  & $O(MK)$ \\[10pt]
         \hline
       $\hat{\nabla}^{(b)}$ & $\frac{\frac{1}{M}\sum_{m=1}^M \hat{C}_{(m)}  f(\tilde z_{(m)})}{\frac{1}{M} \sum_{m=1}^M \hat{C}_{(m)}}$ & $O(M)$ \\[10pt]
         \hline
       $\hat{\nabla}^{(c)}$ & $\frac{\hat{C}_{(M)} \big{(} \sum_{k=1}^K w^k_{(M)} f(z^k_{(M)} \big{)}}{\frac{1}{M} \sum_{m=1}^M \hat{C}_{(m)}}$  & $O(K)$ \\[10pt]
         \hline
         
    \end{tabular}
    
\end{table}

In our amortized setting, because we have a collection of data points $\mathcal{D} = \{x_1, \dots, x_n\}$, our method keeps track of the $M$ different LT-SMC runs for each point $x_j \in \mathcal{D}$. For any point $x_j \in \mathcal{D}$, $\hat{\nabla}_j^{(a)}, \hat{\nabla}_j^{(b)}$ and $\hat{\nabla}_j^{(c)}$ are the three gradient estimators from \Cref{tab:estimators} for a given iteration of the training procedure. The SMC-Wake procedure is detailed \Cref{algorithm:smc_wake}, where we use the generic gradient estimator $\hat{\nabla}$: any of $\hat{\nabla}^{(a)}$, $\hat{\nabla}^{(b)}$, $\hat{\nabla}^{(c)}$ can be used. SMC-Wake interleaves stochastic gradient descent steps that follow the gradient $\frac{1}{n}\sum_{j=1}^n \hat{\nabla}_j$ with LT-SMC runs (\Cref{algorithm:LT-SMC_short}). Here we average over all $n$ training observations, but a mini-batch will also suffice.  Each $\hat{\nabla}_j$ itself is an aggregate of $M$ samplers for each point. Each additional run of LT-SMC computes new particle approximations and normalizing constant estimates for each point $x_j \in \mathcal{D}$.  Each time new particle approximations are computed for a given point $x_j$, the estimators $\hat{\nabla}_j^{(a)}, \hat{\nabla}_j^{(b)}$ or $\hat{\nabla}_j^{(c)}$ are recomputed and the estimate of the normalizing constant $C_j = p(x_j)$ is refined. \Cref{algorithm:smc_wake} suggests rerunning LT-SMC samplers for each point $x_j \in \mathcal{D}$ at each iteration, but SMC-Wake can be made less computationally demanding by distributing LT-SMC runs into batches. In our experiments (\Cref{experiments}), for example, we rerun LT-SMC for only a single observation at each iteration.  It is also possible to decouple the process of running LT-SMC with the process of optimization: perform a large number of LT-SMC runs before optimization begins, and then re-use these particles for every step of the optimization.  
\RestyleAlgo{ruled}
\DontPrintSemicolon
\SetNoFillComment %
\SetKwInput{KwInputs}{Inputs}
\SetKwRepeat{Repeat}{Repeat}{Until Convergence}
\begin{algorithm}[hbt!]
  \caption{\texttt{SMC-Wake}}
  \label{algorithm:smc_wake}
  \KwInputs{Data $x_1, \dots, x_n$, encoder $q_{\phi}$, likelihood $p(x \mid z)$, prior $p(z)$, SGD step size $\eta$, number of particles $K$, temperatures $\tau$. }

    Initialize $\textrm{SMC}_j = \emptyset$, $j=1,\dots, n$\;
    \Repeat{\\}{
    \For{$j=1, \dots, n$ \tcc{(for each datapoint}}{
    $z^{1:K}, w^{1:K}, \hat{C} \gets \texttt{LT-SMC}(x_j, p(x \mid z), p(z), \tau)$ \;
    $\textrm{SMC}_j = \textrm{SMC}_j \cup \{[z^{1:K}, w^{1:K}, \hat{C}]\}$\;
    Compute $\hat{\nabla}_j$ \tcc{(See Table \ref{tab:estimators})}
  }
  $\phi \gets \phi - \eta \frac{1}{n}\sum_{j=1}^n \hat{\nabla}_j$\;
  }
  \textrm{Return $q_\phi$.}\;
\end{algorithm}

\subsection{Asymptotic analysis}
\label{subsection:analysis}

We show that $\hat{\nabla}^{(a)}, \hat{\nabla}^{(b)}$ and $\hat{\nabla}^{(c)}$ are asymptotically unbiased and that the former two estimators are strongly consistent. We begin with a well-known result: the LT-SMC approximations to the posterior become arbitrarily accurate as the number of particles $K \to \infty$.  The following proposition formalizes this result.
\begin{proposition}
  \label{theorem:KL}
  For an observation $x$ and a joint model with prior $p(z)$ and bounded likelihood density $p(x|z)$, let $0 = \tau_1 < \cdots < \tau_T = 1$ denote a tempering schedule. Let $\gamma_{t}(z) \propto p(z)p(x \mid z)^{\tau_{t}}$ denote intermediate targets, and let $M_t(\cdot,dz_t)$ be a sequence of Markov transition kernels that leaves $\gamma_{t-1}$ invariant. Let $\hat{P} \sim \mathrm{Cat}(w_T^{1:K}, z_T^{1:K})$ be the particle distribution produced by LT-SMC using $\{\tau_t, \gamma_t, M_t\}_{t=1}^T$. Then, for any measurable and bounded test function $f$,
  \[
  \lim_{K\rightarrow \infty} \mathbb{E}\left[\left(
  \mathbb{E}_{\hat P}[f(z)] - \mathbb{E}_{p(z \mid x))}[f(z)]
  \right)^2\right] = 0. 
  \]
\end{proposition}

A proof follows by applying Proposition 11.3 of \citet{Chopin2020IntroSMC} with $G_t(z_{t-1},z_t)=p(z_t)p(x|z_t)^{\tau_{t}-\tau_{t-1}}$.
Above, the outer expectation is taken with respect to all variables generated by the SMC algorithm. In our case, we are interested in the test function $f(z) = -\nabla_\phi \log q_\phi(z \mid x)$. \Cref{theorem:KL} implies that gradient estimators based on a single LT-SMC run are asymptotically consistent as $K \to \infty$. In practice, however, finite $K$ must be used,  resulting in biased gradient estimators. 

The estimators $\hat{\nabla}_j^{(a)}, \hat{\nabla}_j^{(b)}$ and $\hat{\nabla}_j^{(c)}$ (\Cref{tab:estimators}) resolve this bias. These estimators with fixed $K$ are asymptotically unbiased as the number of LT-SMC runs $M \to \infty$ and the former two are strongly consistent.

\begin{proposition}
    \label{thm:consistency}
    For an observation $x_j$, suppose that $f(z) = -\nabla_\phi \log q_\phi(z \mid x_j)$ is a bounded and measurable function for any $\phi \in \Phi$. For $m=1,\dots, M$, let the random variables $\hat{C}_{(m)}$,  $w^{1:K}_{(m)}$, and $z^{1:K}_{(m)}$ result from an independent run of LT-SMC (\Cref{algorithm:LT-SMC_short}) with a fixed temperature schedule. Then, the gradient estimators $\hat{\nabla}_j^{(a)}$ and $\hat{\nabla}_j^{(b)}$ are strongly consistent for $\mathbb{E}_{p(z \mid x_j)} f(z)$ in $M$ and the estimator $\hat{\nabla}_j^{(c)}$ is asymptotically unbiased in $M$.
\end{proposition}

\Cref{appendix:consistency} provides a proof. Strongly consistent estimators may be preferred, but are more costly: computing $\hat{\nabla}_j^{(a)}$ and $\hat{\nabla}_j^{(b)}$ require $O(MK)$ and $O(M)$ memory, respectively, for each data point $x_j$. However, in our experiments (\Cref{experiments}), we find that SMC-Wake outperforms wake-phase training even for relatively small $M$, for which memory usage is not a limiting factor, for example $M=100$. Moreover, for any fixed number of particles $K$, the estimator $\hat{\nabla}_j^{(c)}$ is asymptotically unbiased yet requires just $O(K)$ memory (regardless of $M$). This is advantageous compared to wake-phase training, which uses biased gradient estimators and also requires $O(K)$ memory. We further discuss these three estimators and their memory usage in \Cref{appendix:consistency}.

\subsection{Extension to particle MCMC}
\label{subsection:particle_mcmc}

Just as Markov chain Monte Carlo methodology can be incorporated into SMC algorithms (e.g., the mutation kernels in \Cref{algorithm:LT-SMC_short}), SMC samplers can be nested within MCMC frameworks. Particle-Independent Metropolis-Hastings (PIMH) \citep{Andrieu2010ParticleMCMC} is one method of constructing a Markov chain that samples the posterior using multiple runs of LT-SMC. Given a state $z$ and a normalizing constant estimate $\hat{C}$, a proposed draw $z_{\textrm{new}} \sim \hat{P}_{\textrm{new}}$ is accepted with probability $\alpha = \textrm{min}(1, \frac{\hat{C}_{\textrm{new}}}{\hat{C}})$, where $\hat{P}_{\textrm{new}}, \hat{C}_{\textrm{new}}$ are quantities from a new LT-SMC run.

For situations where memory is limiting, we propose adding a PIMH outer loop to SMC-Wake to yield a new algorithm: SMC-PIMH-Wake (\Cref{algorithm:smc_pimh_wake}). Accepting or rejecting new particles within Metropolis-Hastings steps takes the place of iterative aggregation. For any data point $x_j$, this procedure results in a convergent Markov chain with $p(z \mid x_j)$ as its stationary distribution under mild conditions. SMC-PIMH-Wake yields asymptotically unbiased estimates of the gradient of the forward KL divergence as the number of Metropolis-Hastings steps tends to infinity with $O(K)$ memory, similar to SMC-Wake with the gradient estimator $\hat{\nabla}^{(c)}$. As the test function $f(z) = -\nabla_\phi \log q_\phi(z \mid x_j)$ changes with every gradient step, consistent estimates cannot be obtained with finite memory. In situations where the variance of the gradient estimators computed by LT-SMC is large, SMC-PIMH-Wake may provide lower-variance gradient updates, e.g. by using the same SMC sampler to estimate gradients for multiple gradient steps consecutively.

\RestyleAlgo{ruled}
\DontPrintSemicolon
\begin{algorithm}[hbt!]
  \caption{\texttt{SMC-PIMH-Wake}}
  \label{algorithm:smc_pimh_wake}
  Initialize $\hat{P}_j = \textrm{Cat}(z^{1:K}, w^{1:K})$, $\hat{C}_j = \hat{C}$ with $z^{1:K}, w^{1:K}, \hat{C} \gets \texttt{LT-SMC}(x_j, p(x \mid z), p(z), \tau)$\;
  \Repeat{\\}{
  \For{$j=1, \dots, n$}{
    $z_{\textrm{new}}^{1:K}, w_{\textrm{new}}^{1:K}, \hat{C}_{\textrm{new}} \gets \texttt{LT-SMC}(x_j, p(x \mid z), p(z), \tau)$ \;
    Compute $\alpha = \textrm{min}\bigg{(}1, \frac{\hat{C}_{\textrm{new}}}{\hat{C}_j} \bigg{)}$\;
    Set $\hat{P}_j = \textrm{Cat}(z_{\textrm{new}}^{1:K}, w_{\textrm{new}}^{1:K})$, $\hat{C}_j = \hat{C}_{\textrm{new}}$ w.p. $\alpha$ \;
  }
  $\phi \gets \phi - \eta \frac{1}{n}\sum_{j=1}^n \left(-\mathbb{E}_{\hat{P}_j} \nabla_\phi \log q_\phi(z \mid x_j)\right)$\;
  }
\end{algorithm}

\section{RELATED WORK}
\label{related_work}

Variational Sequential Monte Carlo (VSMC) \citep{Naesseth2018VSMC}, Filtering Variational Objectives (FIVO) \citep{Maddison2017FIVO}, and Auto-Encoding Sequential Monte Carlo (AESMC) \citep{Le2018AutoencodingSMC} are closely related to each other; all use SMC to facilitate optimization of bounds on the marginal evidence. In contrast to SMC-Wake, these methods are tailored to sequential data and optimize the reverse KL divergence. Optimizing the reverse KL divergence is particularly difficult within an SMC framework as computing gradients of the reverse KL often relies on the reparameterization trick; however, the resampling step of SMC is not easily reparameterized. By minimizing the forward KL divergence, we eliminate the need for reparameterization. 

Markovian Score Climbing (MSC) \citep{Naesseth2020MarkovianScoreClimbing} is also a method for optimizing the forward KL. It uses an MCMC outer loop to achieve asymptotic guarantees. In this way, MSC differs from SMC-Wake (\Cref{algorithm:smc_wake}), but is similar to SMC-PIMH-Wake.  MSC uses a conditional importance sampling (CIS) transition kernel with $q_\phi$ as the proposal distribution, rather than a Metropolis-Hastings kernel as is used in SMC-PIMH-Wake. We show in \Cref{experiment:gaussian} that our PIMH variant outperforms MSC, which mixes slowly because initially $q_\phi$ is a poor proposal compared to the particles generated by LT-SMC. 

Neural Adaptive Sequential Monte Carlo (NASMC) \citep{Gu2015NeuralAdaptiveSMC} is a non-amortized method for sequential data (e.g., state-space models) that adapts SMC mutation kernels to minimize the forward KL divergence to the intermediate targets. SMC-Wake differs from NASMC in its use of Metropolis-Hastings kernels that do not depend on $q_\phi$, and in its gradient calculations that only use particle approximations to the final target $\gamma_T$. Like RWS, NASMC gradient estimators are also biased for finite $K$. Three related methods, namely, Annealed Flow Transport (AFT) \citep{Arbel2021AFT}, Continual Repeated Annealed Flow Transport \citep{Matthews22CRAFT}, and Nested Variational Inference (NVI) \citep{Zimmerman2021NestedVariationalInference}, also sample from a target distribution by fitting proposals to minimize a sequence of divergences to intermediate targets. These sampling methods can be incorporated into the SMC-Wake framework: while we propose using LT-SMC samplers, any SMC-based algorithm can be used. The circular pathology is avoided as long as the parameters of the proposals are not shared with the amortized encoder $q_\phi$. Using methods such as these in place of the LT-SMC subroutine in \Cref{algorithm:smc_wake} is a potential direction for future research.

\section{EXPERIMENTS}
\label{experiments}

\subsection{Two moons}
\label{experiments:two_moons}

The ``two moons'' model has been extensively used in the simulation-based inference literature to benchmark existing algorithms \citep{Greenberg2019AutoPosteriorTransformation, Lueckmann2021BenchmarkingSBI}. The generative process for latent $z \in \mathbb{R}^2$ and observed $x \in \mathbb{R}^2$ first draws $z_1, z_2 \overset{iid}{\sim} U(-1,1)$, and then draws two auxiliary variables $a \sim U\left(-\frac{\pi}{2}, \frac{\pi}{2}\right)$ and $r \sim \mathcal{N}\left(0.1,0.01^2\right)$. 
Finally, the observation
\begin{align*}
x^\top & =p+\left[-\frac{\left|z_1+z_2\right|}{\sqrt{2}}, \frac{-z_1+z_2}{\sqrt{2}}\right],
\end{align*}
where $p  =[r \cos (a)+0.25, r \sin (a)]$. The posterior distribution on $z$ given $x$ takes the shape of two crescent moons facing each other. Because these regions are disconnected, separated by a zero density region of the posterior, exploring the latent space is potentially difficult and a highly flexible variational family is required to approximate the posterior well.  We use the neural spline flow (NSF) as the variational family, conditional on a given observation $x$ \citep{Durkan2019NeuralSplineFlow}. 

Given 100 points $x_1, \dots, x_{100}$ generated independently from the two-moons model, we compare SMC-Wake with estimator $\hat{\nabla}^{(a)}$, RWS (wake-phase training only), and a defensive variant of RWS that performs importance sampling using the mixture $\frac{1}{2}p(z) + \frac{1}{2}q_\phi(z \mid x)$ as a proposal. Defensive importance sampling has been found to ameliorate the mass concentration problem that can occur in wake-phase training when $q_\phi$ is used as the proposal \citep{Le2019RevisitRWS}. Additionally, this choice of proposal avoids a divide-by-zero error that arises when the proposed particles all have zero as their posterior densities.

Additional details of the training procedure and results for $\hat{\nabla}^{(b)}$ and $\hat{\nabla}^{(c)}$ are in \Cref{appendix:two_moons}. \Cref{fig:moon_plots} shows the amortized variational posteriors $q_\phi(z \mid x_{16})$ as an example. SMC-Wake is the only method that captures the shape of the posterior. Wake-phase training severely concentrates mass into a single point-like region, exhibiting the degeneracy described in \Cref{section:mass_concentration}. \Cref{appendix:two_moons} shows that problematic mass concentration for RWS occurs quickly, within 1000 gradient steps. While the defensive variant of RWS does better, it is still far worse than the quality of the SMC-Wake approximations. Variational approximations for several additional data points are shown in \Cref{appendix:two_moons}.

\begin{figure}[ht!]
  \centering
  \includegraphics[height=190pt, width=.9\linewidth]{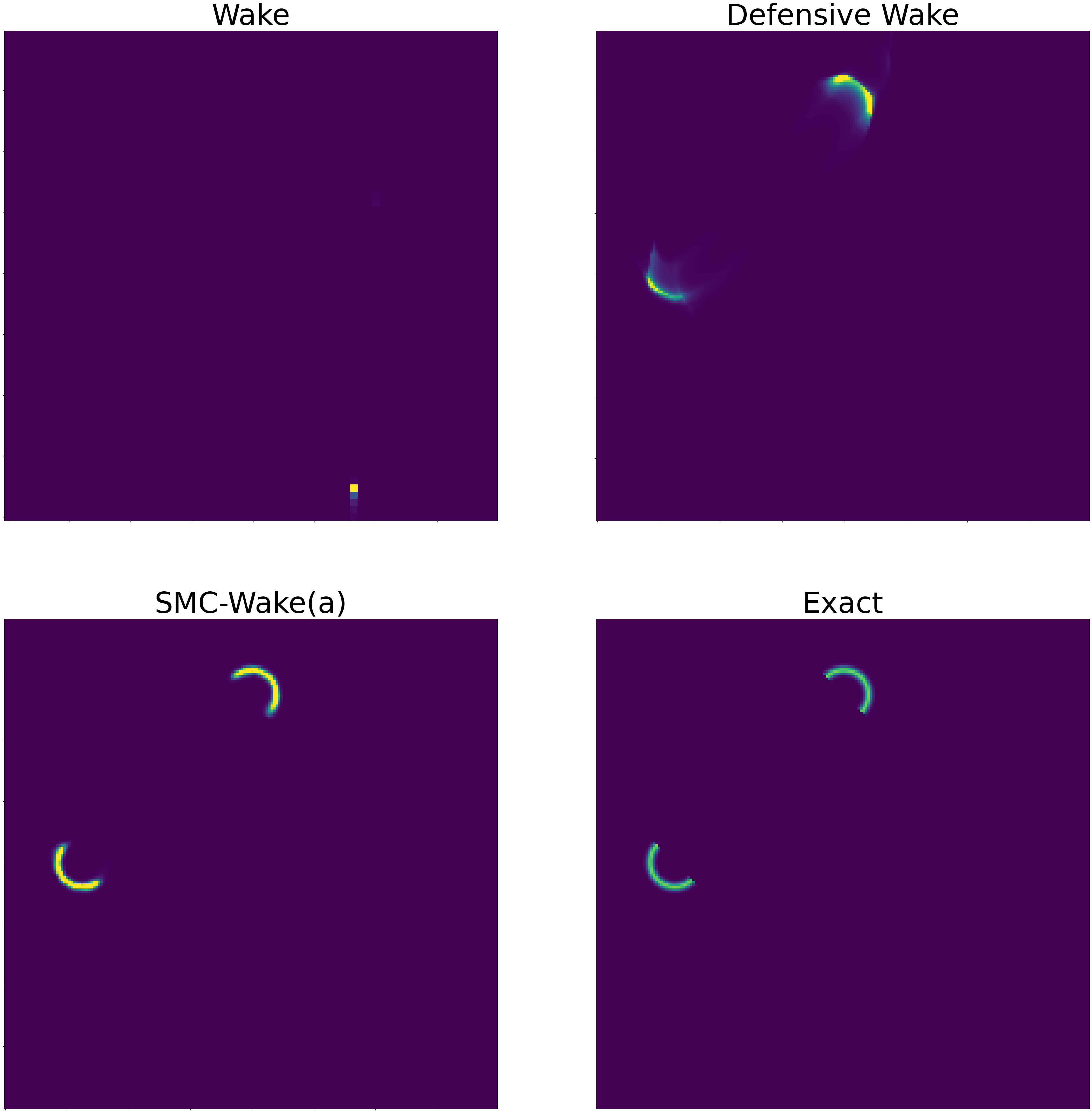}
  \caption{Posterior approximations given $x_{16}$ by SMC-Wake, Wake, and Defensive Wake. The bottom right panel depicts the exact posterior distribution.}
\label{fig:moon_plots}
\end{figure}

\subsection{Avoiding mode collapse in MNIST}
\label{experiment:normalized_mnist}

We consider learning a model of MNIST digits to illustrate that SMC-Wake, though an inference method, can nonetheless aid in model learning. Given 1000 normalized MNIST digits and labels $\{x_i, \ell_i\}_{i=1}^{1000}$, we fit a conditional model $p(x \mid \ell, z)$ to maximize the importance-weighted bound (IWBO) (\Cref{appendix:elbo_iwbo}) while simultaneously fitting the encoder $q_\phi$. This follows the reweighted wake-sleep (RWS) framework of \citet{BornscheinBengio2015RWS}. The model is based on a sigmoid belief network \citep{Saul1996SigmoidBeliefNetwork} and is given by
\begin{equation}
    p_\psi(x \mid \ell, z) \sim \mathcal{N}(\sigma\big{(}W_\ell z + b_\ell\big{)}, \tau^2 I_d),
\end{equation}
where $\sigma(\cdot)$ denotes the sigmoid function, $\ell \in \{0,\dots,9\}$ denotes the label, and $\psi = \{W_0,\dots, W_9, b_0,\dots, b_9\}$ denotes model parameters to be learned. The hyperparameter $\tau=0.01$ results in a highly peaked likelihood. As the labels $\ell$ are fixed, this model fitting task is equivalent to fitting a separate model for each digit class. The amortized encoder $q_\phi(z \mid x, \ell)$ is fit jointly with the model using both SMC-Wake and wake-phase training, with alternating gradient updates for $\phi$ and $\psi$, respectively. We use the gradient estimator $\hat{\nabla}^{(b)}$ in this example.

\begin{figure}[ht!]
\centering
\includegraphics[height=3.5cm,width=\linewidth]{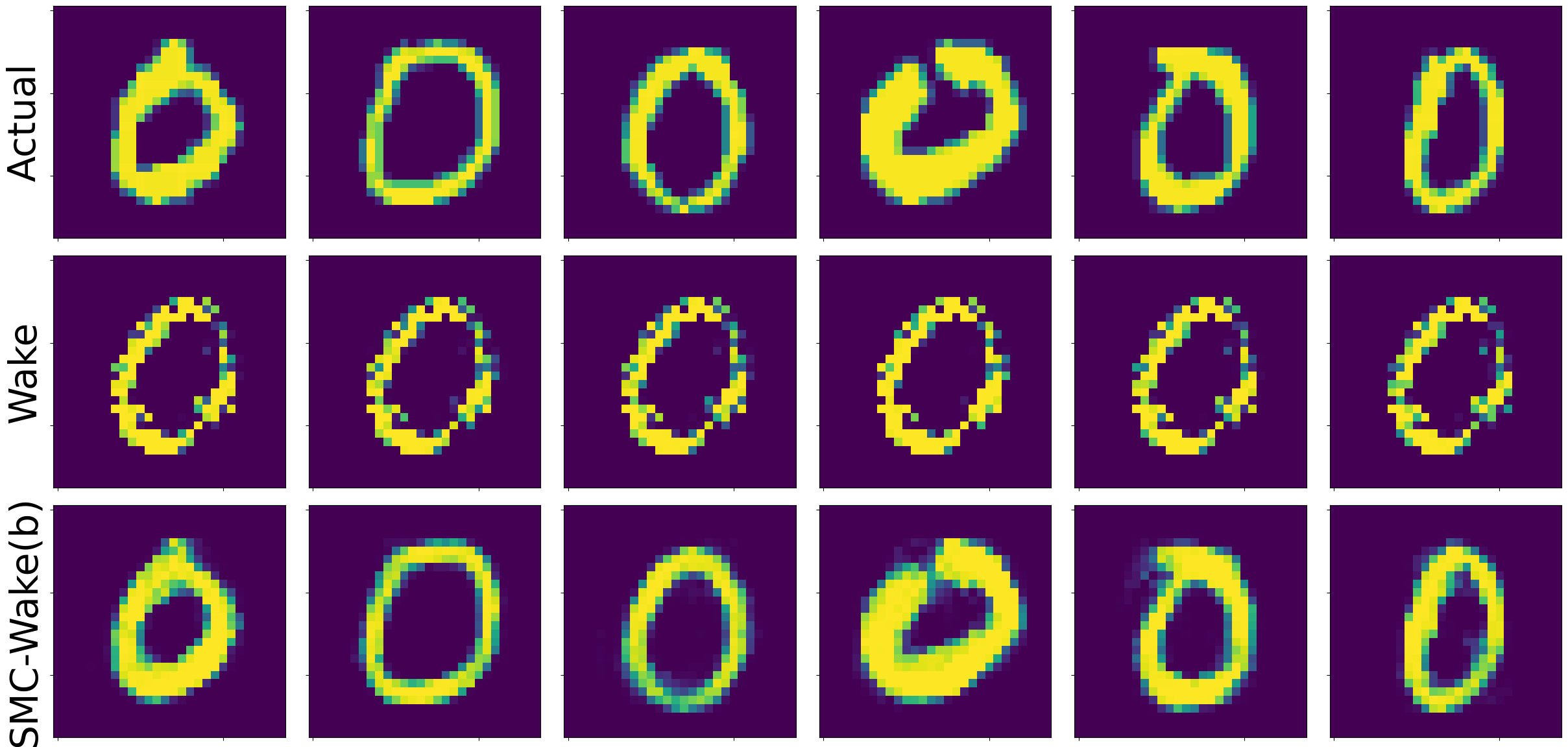}
\caption{Illustration of mass concentration in wake-phase training. Wake-phase reconstructions (middle) of real MNIST digits (top) with label zero all look nearly identical, unlike SMC-Wake reconstructions (bottom).}
\label{fig:mnist_reconstructions}
\end{figure}

Additional details of the implementation are given in \Cref{appendix:mnist}. Wake-phase gradient updates fit an encoder network that severely concentrates mass, learning similar latent representations $z$ for all digits in a given class $\ell$. This results in mode collapse in the generative model, visible by visualizing reconstructions (\Cref{fig:mnist_reconstructions}). Given different instances of the zero digit class, for example, the wake-phase reconstructions (middle) look nearly identical. This occurs due to the pathological concentration of $q_\phi(z \mid x, \ell)$ described in \Cref{section:mass_concentration}, resulting in nearly identical latent representations for all instances of a digit $\ell$. \Cref{appendix:mnist} shows similar behavior for the other digit classes as well. SMC-Wake avoids this degeneracy; its reconstructions (bottom) closely match the true images (top).

\subsection{Transformed Gaussian} 
\label{experiment:gaussian}

\subsubsection{Nested MCMC approaches}

We consider a Gaussian-Gaussian hierarchical model with latent variable $z \sim \mathcal{N}(0, \sigma^2 I_p)$ and data $x \mid z \sim \mathcal{N}(Az, \tau^2 I_d)$ to compare two methods that make use of an MCMC outer loop: SMC-PIMH-Wake (\Cref{subsection:particle_mcmc}) and Markovian Score Climbing (MSC) \citep{Naesseth2020MarkovianScoreClimbing}. MSC proposes particles for the next step of the Markov chain using $q_\phi$. If this proposal is poor, then the Markov chain mixes slowly, and thus may concentrate mass. We study a high-dimensional case with $p=50$ and $d=100$, and take standard deviations $\sigma = \tau = 1$. The design matrix $A \in \mathbb{R}^{d \times p}$ is full-rank and fixed. 

Both methods fit a variational family $q_\phi(z \mid x) \sim \mathcal{N}(\mu, L L^\top + \epsilon I)$ with a lower-triangular matrix $L \in \mathbb{R}^{p \times p}$. This family is flexible enough to approximate the posterior almost exactly. (The addition of $\epsilon I$ with $\epsilon = 0.0001$ provides numerical stability.) The variational posterior amortizes over $n=50$ points $\{x_j\}_{j=1}^{50}$ generated from the above model. The exact posterior distribution can be written in closed form by completing the square (see \Cref{appendix:gaussian}); this enables analytical computation of the forward, reverse, and symmetric KL divergences for comparison. All divergences are averaged over the $n=50$ observations.

 We used $K=100$ particles for importance sampling in MSC and $K=100$ particles for LT-SMC in SMC-PIMH-Wake. We find that MSC mixes slowly, and thus performs poorly compared to SMC-PIMH-Wake. \Cref{tab:gaussian_gaussian_results} presents results after 500,000 gradient steps for MSC, corresponding to at least 250,000 MCMC steps for each point $x_j \in \mathcal{D}$ with mini-batching. In contrast, the results for SMC-PIMH-Wake are computed after only 40,000 gradient steps. As LT-SMC is more expensive than importance sampling, we perform fewer MCMC steps, only 800 MCMC steps for each $x_j$. (\Cref{appendix:gaussian} provides additional details about our implementation.) SMC-PIMH-Wake leads to lower average forward KL divergence than MSC. The mixing time for MSC is large because initially $q_\phi$ is a poor proposal, whereas from the first iteration SMC-PIMH-Wake proposes from a high-quality approximation to the posterior computed by LT-SMC. 
\begin{table}[ht!]
    \caption{Gaussian Hierarchical Model Learning Results (Lower is better.)}
    \centering
\begin{tabular}{lrr}
\toprule
 & MSC & SMC-PIMH-Wake \\
\midrule
Forward KL & 2339 & \textbf{1387} \\
Reverse KL & 1762 & \textbf{1287} \\
Symmetric KL & 4102 & \textbf{2674} \\
\bottomrule
\end{tabular}
\label{tab:gaussian_gaussian_results}
\end{table}
\raggedbottom

\subsubsection{The importance of averaging}
\begin{figure}[ht!]
    \centering
    \includegraphics[width=\linewidth]{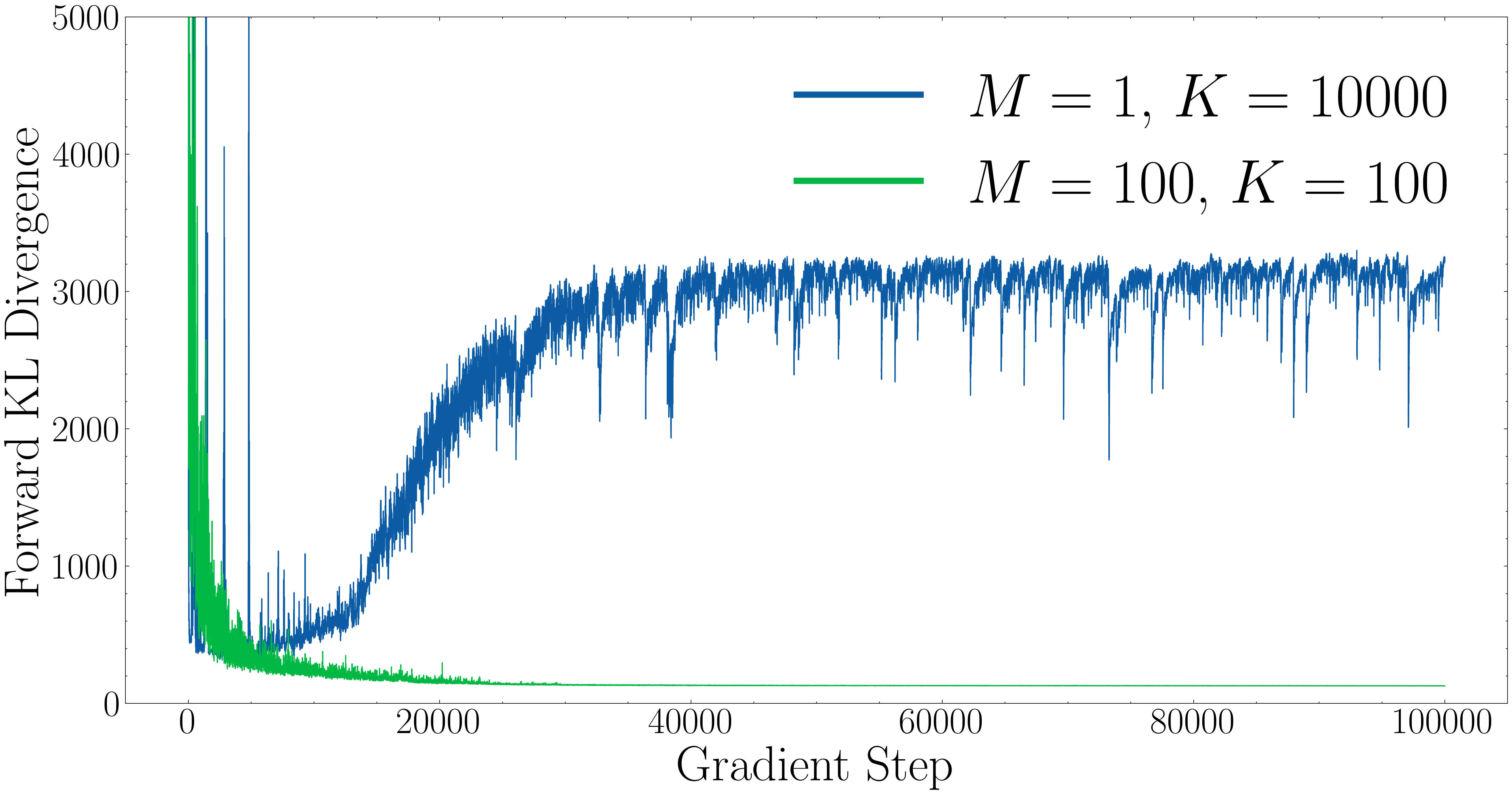}
    \caption{Average forward KL divergence during training for an encoder $q_\phi$ fit using one $K=10,000$ sampler per-point (blue) and $M=100$ different $K=100$ samplers (green).}
    \label{fig:many_samplers_vs_one}
\end{figure}

We now illustrate that for a fixed computational budget, the approach used in SMC-Wake of taking a small $K$ and using many different samplers outperforms the naive approach of taking large $K$ and using a single sampler. We use a similar setting as above. (For implementation details, see \Cref{appendix:gaussian}.) For each observation, we run LT-SMC with $K=10,000$ particles and also perform $M=100$ runs each using $K=100$ particles. Gradient estimators are constructed from each of these two settings and used to fit $q_\phi$. We use a naive estimator for the $M=100$ case that does not weight according to $\hat{C}_{(m)}$ values. Both settings require a similar number of likelihood evaluations. \Cref{fig:many_samplers_vs_one} demonstrates that fitting the encoder with $M=100$ samplers results in a $q_\phi$ with significantly lower average forward KL divergence to the posterior. 

The larger $K$ counterintuitively performs worse because it suffers from a lack of sample replenishment \citep{Matthews22CRAFT}. If the $K=10,000$ SMC sampler has a low effective sample size (ESS), as occurs in this case due to the large dimension $p=50$ of the latent space, the encoder $q_\phi$ is significantly overfit to only a few particles. Although these particles may be located at a peak of the posterior density, the lack of diversity among them hinders the fitting process. In contrast, even if all $M=100$ samplers of $K=100$ have low $\mathrm{ESS}$, the diversity of samples is richer and averaging over many independent samplers reduces variance (at the cost of increased bias due to smaller $K$).

\subsection{Galaxy spectra emulator}
The Probabilistic Value-Added Bright Galaxy Survey (PROVABGS) simulator is a state-of-the-art simulator of astronomical spectra \citep{Hahn2022Provabgs} that maps parameters $\theta$ to spectra $x$. Each spectrum is a high-dimensional vector of flux measurements for an astronomical object: at each wavelength $\lambda$ in a grid between 3000 and 10,000 angstroms, the spectrum records the flux, typically measured in units $\textrm{erg} \cdot \textrm{cm}^{-2} \textrm{s}^{-1} \textrm{{\AA}}^{-1}$ \citep{York2000SDSSTechnicalSummary, Abareshi2022DESIOverview}.  We consider the problem of predicting parameters from observed spectra: given noisy samples of the spectra $x$, we will learn to reconstruct the astronomical parameters $\theta$.

We train a neural network emulator of the PROVABGS simulator for fast likelihood evaluation and work with normalized spectra for ease. Details are given in \Cref{appendix:sed}. Our emulator takes 11 scientific input parameters $\theta$ of interest as inputs and produces spectra as outputs. (Examples are given in \Cref{appendix:sed}.) We use the priors for the 11 parameters suggested by \citet{Hahn2022Provabgs}. As the simulator's output is noiseless, we add Gaussian noise proportional to the signal strength in each dimension to simulate typical measurement error, with signal-to-noise ratio (SNR) set to $1/\sigma$ with $\sigma = 0.1$. The likelihood function is thus multivariate Gaussian and can be evaluated efficiently. 

\begin{figure}[ht!]
  \centering
    \includegraphics[height=200pt,width=.99\linewidth]{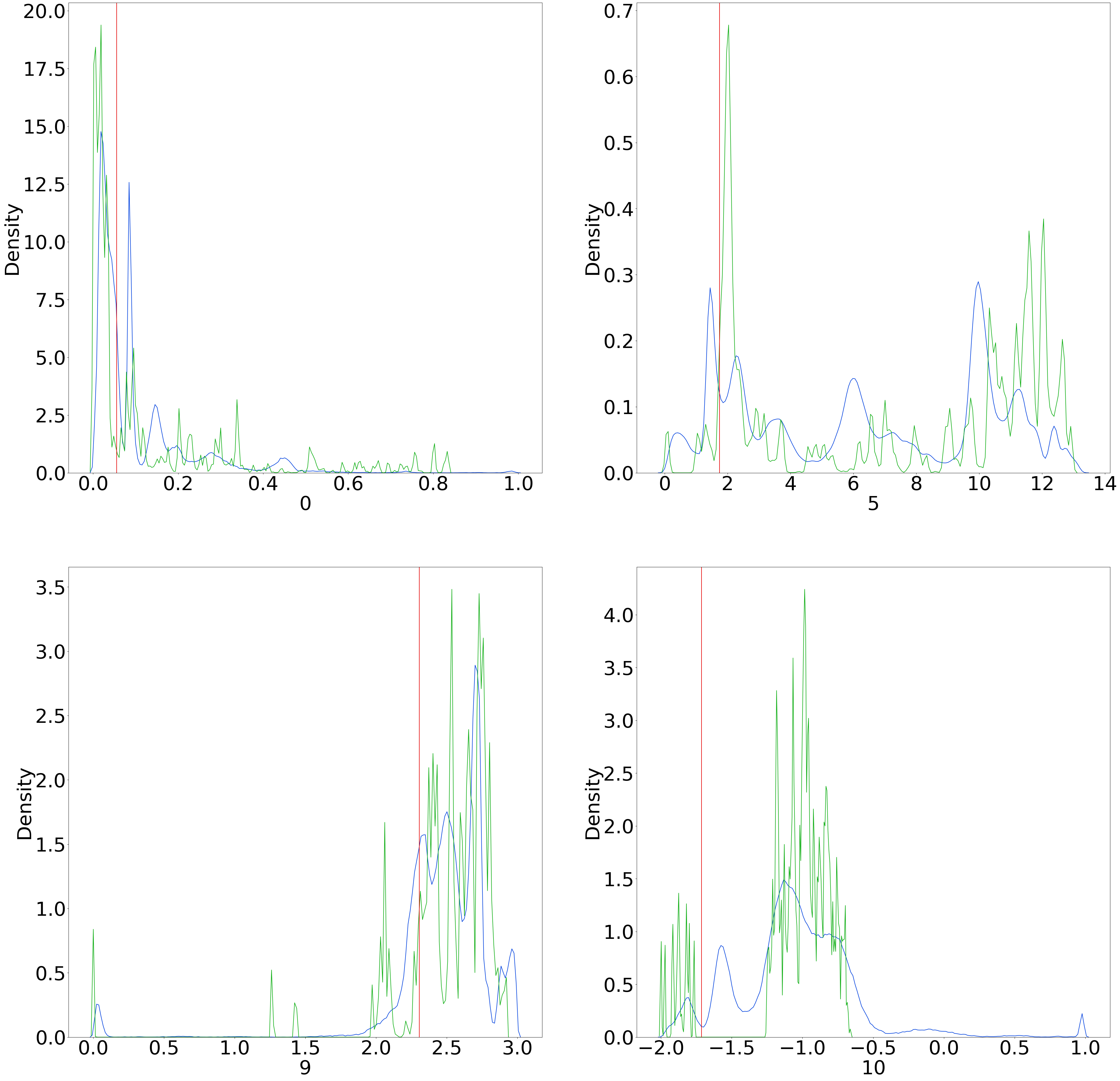}
    \caption{Posterior estimation for one example spectrum $x$ from the training set.  We used MCMC (green) and SMC-Wake (blue) to estimate the posterior on $\theta$.  Smoothed (marginal) estimates of the posterior for 4 of the 11 parameters $\theta_i$ are plotted, with the true value of the parameter indicated with a red vertical line.}
    \label{fig:smcwake_mcmc} 
  \end{figure}

Given 100 spectra from the forward model, we fit an amortized neural spline flow $q_\phi(\theta \mid x)$ using SMC-Wake with estimator $\hat{\nabla}^{(a)}$ and $K=100$ particles. Additional implementation details are given in \Cref{appendix:sed}. To assess the quality of the amortized encoder, as a surrogate for the ground truth we run MCMC using Metropolis-Hastings random walks with 100 walkers for 10,000 steps each following 10,000 burn-in steps. We plot several kernel-smoothed marginal density estimates from both MCMC and SMC-Wake in \Cref{fig:smcwake_mcmc}. SMC-Wake generally captures the shape of the posterior well, assigning high density to the ground truth, while being somewhat overdispersed as a consequence of minimizing the forward KL divergence. This compares favorably to wake-phase training, which exhibits numerical instability on this problem (\Cref{appendix:sed}).

\section{DISCUSSION}

SMC-Wake is a method for minimizing the forward KL divergence in amortized variational inference using likelihood-tempered SMC. SMC-Wake generalizes wake-phase training from Reweighted Wake-Sleep (RWS) and improves on it by providing consistent or asymptotically unbiased gradient estimates for the forward KL divergence. Wake-phase updates have a troublesome circular property: $q_\phi$ is used to make proposals that are themselves used to update $\phi$.  Empirical results demonstrate that this entanglement can lead to a degenerate concentration of mass. SMC-Wake avoids this degeneracy by proposing from the prior. Exploring ways to safely incorporate $q_\phi(z \mid x)$ into the proposal step is an interesting direction for future research.

SMC-Wake may not be appropriate in all settings.  Each run of SMC is more expensive than the lightweight importance sampling used by wake-phase updates.  However, in practice, we find that SMC yields more accurate answers. With further research, we speculate that incorporating proposal kernels based on likelihood gradients, such Metropolis-Adjusted Langevin kernels \citep{Roberts1998MetroAdjustLangevin} into SMC-Wake may lead to even better performance.

\clearpage
\section*{Acknowledgements}
 We thank the reviewers for their helpful comments and suggestions. The authors are grateful for support from the National Science Foundation under grant numbers 2209720 (OAC) and 1841052 (DGE).

\printbibliography

\clearpage
\appendix

\onecolumn
\section{OBJECTIVES FOR AMORTIZED INFERENCE}
\label{appendix:elbo_iwbo}

Given a data point $x$, variational inference seeks to approximate the intractable posterior $p(z \mid x)$ with a tractable distribution $q(z)$ \citep{Blei2017ReviewVI,Jordan2008VI}. Learning a different $q$ for every $x$ of interest can be computationally expensive. Amortized variational inference (AVI) leverages deep learning for this task, learning parameters $\phi$ of a neural network that transforms data $x$ to the distributional parameters of a posterior approximation. AVI thus allows us to create the conditional distribution $q_\phi(z \mid x)$ for any $x$ using a forward pass through this neural network \citep{Ranganath2014Blackbox,Ambrogioni2019FAVI,Kingma2019VAEIntro}. In many settings, the decoder $p_\psi(x \mid z)$ is also learned along with the inference network, with both fit to an observed dataset $\mathcal{D} = \{x_1,\dots, x_n\}$.

To learn the model parameters $\psi$ and the variational parameters $\phi$, variational autoencoders (VAEs) \citep{Kingma2019VAEIntro} target the reverse KL divergence $\textrm{KL}(q_\phi(z \mid x) \mid \mid p_\psi(z \mid x))$ via maximization of 
\begin{equation}
\label{elbo}
    \textrm{ELBO}(\psi, \phi) = \mathbb{E}_{z \sim q_\phi(z \mid x)} \log \frac{p_\psi(x,z)}{q_\phi(z \mid x)}.
\end{equation}
The importance-weighted autoencoder (IWAE) targets a similar but tighter lower bound on the evidence, a multi-sample generalization of the ELBO called the importance-weighted bound (IWBO):
\begin{equation}
\label{iwbo}
    \textrm{IWBO}(\psi,\phi) = \mathbb{E}_{z_1, \dots, z_K \overset{iid}{\sim} q_\phi(z \mid x)} \log \sum_{i=1}^K \frac{1}{K} \frac{p_\psi(x,z_i)}{q_\phi(z_i \mid x)}.
\end{equation}
Both ELBO and IWBO maximization can be viewed as the minimization of a reverse KL divergence. In the case of the ELBO, maximization is equivalent to minimization of $\textrm{KL}(q_\phi(z \mid x) \mid \mid p_\psi(z \mid x))$, while for the IWBO, the divergence is between two distributions on an extended sampling space \citep{Kingma2019VAEIntro, Le2018AutoencodingSMC}. 

To achieve lower variance gradients with respect to $\phi$, implementations of both the IWBO and ELBO objectives often rely on the ``reparameterization trick'', writing a draw $z \sim q_\phi(z \mid x)$ as a deterministic function of an auxiliary random variable that is independent of $\phi$ \citep{Kingma2019VAEIntro,Burda2016IWAE}. This reliance on reparameterization stems from the fact that expectations are computed with respect to draws from $q$, which themselves depend on $\phi$. On the contrary, Reweighted Wake-Sleep and any other method targeting the forward KL divergence need not use reparameterization, as expectations in the loss function are computed with respect to $p$ (which is constant with respect to $\phi$) rather than $q_\phi$.

\section{THE CIRCULAR PATHOLOGY}
\label{appendix:circular}

We provide more detail on the surrogate objective (\autoref{eqn:surrogate_objective}). Let $\bot$ denote a stop gradient operator \citep{Scibior2021Stopgrad}. Note that $\bot$ does not affect evaluation whatsoever; $f(x) = f(\bot x)$ always. However, the $\bot$ notation does alter gradient calculation, i.e. $\bot x$ is always considered as a constant with respect to any gradient operation. Stop gradients are necessary when considering the surrogate objective for wake-phase training because they accurately describe the procedure for implementation of wake-phase training in practice. More precisely, to compute
\begin{equation}
    -\sum_{i=1}^K w^i \nabla_\phi \log q_\phi(z^i \mid x)
\end{equation}
one first a) draws $z^{1:K} \overset{iid}{\sim} q_\phi$ and disables gradient tracking with respect to $\phi$ (e.g., by calling \texttt{.sample()} instead of \texttt{.rsample()} in PyTorch), and secondly b) ensures the normalized weights do not track gradients either by calling \texttt{weights.detach()} or similar. In this fashion, automatic differentiation and back propagation operations function as required, and track correct gradients for wake-phase training. Symbolically, the surrogate objective achieves this same behavior with appropriate stop-gradients, as given below:
\begin{equation*}
    \mathbb{E}_{z^1,\dots,z^K \overset{iid}{\sim} q_{\bot \phi}(z \mid x)} -\sum_{i=1}^K (\bot w^i) \log q_\phi(z^i \mid x).
\end{equation*}

We can verify that the Monte Carlo gradient of surrogate objective explicitly matches the Monte Carlo gradient estimator used by wake-phase training. We have
\begin{align}
   \nabla_\phi \mathbb{E}_{z^{1:K} \overset{iid}{\sim} q_{\bot \phi}(z \mid x)} \sum_{i=1}^K(\bot w^i) \log q_\phi(z^i \mid x) &= \nabla_\phi \int  \left(\sum_{i=1}^K(\bot w^i) \log q_\phi(z^i \mid x)\right)q_{\bot \phi}(z^{1:K} \mid x) dz^{1:K} \\
   &= \int  \left(\sum_{i=1}^K(\bot w^i) \nabla_\phi \log q_\phi(z^i \mid x)\right)q_{\bot \phi}(z^{1:K} \mid x) dz^{1:K} \\
   &=  \mathbb{E}_{z^{1:K} \overset{iid}{\sim} q_{\bot \phi}(z \mid x)} \sum_{i=1}^K(\bot w^i) \nabla_\phi \log q_\phi(z^i \mid x) 
\end{align}
where we just used the basic properties of $\bot$ in the derivation above, namely quantities with $\bot$ are constant with respect to the gradient operator $\nabla_\phi$. A Monte Carlo draw of the final line is equivalent to the wake-phase SNIS gradient estimator with $K$ particles. Accordingly, wake-phase dynamics in $\phi$ follow the gradient of this surrogate objective (with the appropriate stop gradient operations in place). 

We next prove \Cref{prop:formal_circular}, reproduced below.

\begin{customprop}{1}
    Let $\mathcal{L}(q)$ denote the surrogate objective defined above for fixed $x$ and fixed $K \in \mathbb{N}$. Let $p$ denote the posterior $p(z \mid x)$. Then there exists $q(z) \neq p(z \mid x)$ such that $\mathcal{L}(q) < \mathcal{L}(p)$.
\end{customprop}

\begin{proof}
     Fix $c \in \mathbb{R}$. For any $\delta > 0$, consider $\mathcal L(q_\delta)$ where $q_\delta \sim \mathrm{Unif}(0,\delta)$. Then, there exists $\delta > 0$ such that $\mathcal L(q_\delta) < c$. Because the density of $q$ is given by $\frac{1}{\delta}$, we must have $\mathcal L(q_{\delta}) = -\log (1/\delta)$ because the normalized weights sum to one. For any choice of $c \in \mathbb{R}$, provided $\delta < \exp(c)$, we have $L(q_\delta) < c$ as desired. Clearly, then, for sufficiently small $\delta$ we have $\mathcal{L}(q_\delta) < \mathcal{L}(p)$.
\end{proof}

\section{SMC-WAKE IMPLEMENTATION}
\label{appendix:algo}

\Cref{algorithm:LT-SMC_short} below sketches LT-SMC with a fixed temperature schedule. Fixed temperature scheduling is necessary technically for our proofs of asymptotic results to hold. However, either adaptive or non-adapative selection can be used in practice generally: one practical way of exploiting adaptive selection in the fixed temperature framework is to perform exploratory runs to determine a suitable temperature schedule prior to fitting of the encoder network, and thereafter using this fixed schedule throughout.

\RestyleAlgo{ruled}
\DontPrintSemicolon
\SetKwInput{KwInputs}{Inputs}
\SetKwRepeat{Repeat}{Repeat}{Until Convergence}
\begin{algorithm}[hbt!]
\caption{\texttt{LT-SMC}}
\label{algorithm:LT-SMC_short}
\KwInputs{Data point $x$, likelihood function $p(x \mid z)$, prior $p(z)$, temperatures $\tau$.}
Initialize particles $z_1^{1:K} \overset{iid}{\sim} p(z)$, unnormalized weights $\Tilde{w}_1^{1:K} = 1$, and normalized weights $w_1^{1:K} = \frac{1}{K}$.\;
\For{$t=1$ \KwTo $T-1$}
{   
Resample $z_{t}^{(i)} \sim \overset{iid}{\sim} \textrm{Cat}(z_{t}^{1:K}, w_t^{1:K}), \ i \in [K]$ \;
Mutate $z^i_{t+1} \sim M(z^{(i)}_{t}, dz_{t+1}), \ i \in [K]$\;
Update $\tilde{w}_{t+1}^{i} = p(x \mid z_{t+1}^i)^{\tau_{t+1}-\tau_t}, \ i \in [K]$ \;
Normalize $w_{t+1}^i = \frac{\tilde{w}_{t+1}^i}{\sum_{j=1}^K \tilde{w}_{t+1}^j}, \ i \in [K]$ \;
}
Return $z_T^{1:K}, w_T^{1:K}$.
\end{algorithm}

In \Cref{algorithm:LikelihoodTemperedSMC}, we present the likelihood-tempered SMC procedure that is used in our implementations of SMC-Wake for \Cref{experiments}\footnote{Code to reproduce experimental results is publicly available at \texttt{https://github.com/declanmcnamara/smc-wake}}. The algorithm is based on Algorithms 17.1 and 17.3 of \citep{Chopin2020IntroSMC}. The algorithm takes a data point $x$ along with a prior density $p(z)$ and a conditional likelihood $p(x \mid z)$ as input. The transition kernel $M(\cdot, dz)$ at stage $t$ leaves invariant the marginal distribution on $z$ with density proportional to $p(z)p(x \mid z)^{\tau_t}$ by constructing a Metropolis-Hastings random-walk kernel. Temperatures are chosen adaptively via \Cref{algorithm:AdaptiveTempering}, as discussed below. The primary differences between the implementation and the abbreviated \Cref{algorithm:LT-SMC_short} are the use of adaptive temperature selection and the optional resampling step, which can reduce variance. 

\begin{algorithm}[hbt!]
\caption{\texttt{LikelihoodTemperedSMC} \citep{Chopin2020IntroSMC}}
\label{algorithm:LikelihoodTemperedSMC}
\KwInputs{Data point $x$, likelihood function $p(x \mid z)$, prior $p(z)$, invariant kernel $M(\cdot, dz)$, number of particles $K$, minimum effective sample size $\textrm{ESS}_{\textrm{min}}$.}
Initialize stage counter $t=1$, particles $z_1^{1:K} \overset{iid}{\sim} p(z)$, unnormalized weights $\tilde{w}_1^{1:K} = 1$, normalized weights $w_1^{1:K} = \frac{1}{K}$, and $\tau_1 = 0$\;
\For{$t=1$ \KwTo $T-1$}
{
 $\delta \gets \texttt{AdaptiveTempering}(\tau, p(z), p(x \mid z), \{z^{1:K}_t, w^{1:K}_t\})$ \;
  $\tau_{t+1} \gets \tau_t + \delta$\;
  \uIf{\upshape $\textrm{ESS} < \textrm{ESS}_{\textrm{min}}$}{
    Resample $z_{t}^{(i)} \overset{iid}{\sim} \textrm{Cat}(z_{t}^{1:K}, w_t^{1:K}), \ i \in [K]$ \;
    $\hat{w}^{1:K}_t \gets 1$\;
  }
  \uElse{$z_{t}^{(i)} = z_t^i,  \ \ i \in [K]$\; $\hat{w}^{1:K}_t \gets w^{1:K}_t$}
Mutate $z^i_{t+1} \sim M(z^{(i)}_{t}, dz_{t+1}), \ i \in [K]$\;
Update $\tilde{w}_{t+1}^{i} = \hat{w}_{t}^i \cdot p(x \mid z_{t+1}^i)^{\tau_{t+1}-\tau_t}, \ i \in [K]$\;
Normalize $w_{t+1}^i = \frac{\tilde{w}_{t+1}^i}{\sum_{j=1}^K \tilde{w}_{t+1}^j}, \ i \in [K]$\;
}
\textrm{Return $\hat{P} = \textrm{Cat}(z^{1:K}_T, w^{1:K}_T)$.}\;
\end{algorithm}

\Cref{algorithm:AdaptiveTempering} presents the adaptive tempering algorithm featured in \Cref{algorithm:LikelihoodTemperedSMC}, reproduced from \citet{Chopin2020IntroSMC} \S 17.2.3. The notation featured in the algorithm comes from writing the target distribution in energy form \citep{Murphy2023ProbML}, i.e. $\gamma(z) \propto \frac{1}{L} p(z) \exp (-V(z))$, which can be easily achieved, for example with the posterior, by taking $V(z) = -\log p(x \mid z)$. 

\begin{algorithm}[hbt!]
  \caption{\texttt{AdaptiveTempering} \citep{Chopin2020IntroSMC}}
  \label{algorithm:AdaptiveTempering}
  \KwInputs{Current temperature $\tau$, likelihood function $p(x \mid z)$, current particle set $\{z^{1:K}, w^{1:K}\}$, minimum effective sample size $\textrm{ESS}_{\textrm{min}}$.}

  Find $\delta$ that solves \[\frac{\left\{\sum_{i=1}^K \exp \left\{-\delta V\left(z^i\right)\right\}\right\}^2}{\sum_{i=1}^K \exp \left\{-2 \delta V\left(z^i\right)\right\}}=\mathrm{ESS}_{\min },\]
  where $V(z_i) = -\log p(x \mid z^i)$. \;

  \textrm{Return $\delta$.}\;
\end{algorithm}

In practice, we find it unnecessary and computationally expensive to increment the number of SMC sampler runs $M$ for every $x_j \in \mathcal{D}$ each time a gradient step is taken, as is suggested by \Cref{algorithm:smc_wake}. This would require $n$ different LT-SMC runs for every gradient step. As elaborated below in our experimental details, we find it sufficient to increment $M$ for only a single $x_j \in \mathcal{D}$ every gradient step, or perhaps even less often, e.g. every ten gradient steps. 

One consequence of this computation-saving scheme is that the number of LT-SMC runs $M_j$ is different for each observation $x_j$. This complicates vectorized operations when averaging over mini-batches of $x_j \in \mathcal{D}$. To resolve this, our implementation of the gradient estimators $\hat{\nabla}^{(a)}$ and $\hat{\nabla}^{(b)}$ uses a slight alteration of the estimators in \Cref{tab:estimators}. From the equalities
\begin{align*}
    \hat{\nabla}^{(a)} &= \sum_{m=1}^M \frac{\hat{C}_{(m)}}{\sum_{m=1}^M \hat{C}_{(m)}} \big{(} \sum_{k=1}^K w^k_{(m)} f(z^k_{(m)}) \big{)}
\end{align*}
and
\begin{align*}
    \hat{\nabla}^{(b)} &= \sum_{m=1}^M \frac{\hat{C}_{(m)}}{\sum_{m=1}^M \hat{C}_{(m)}}  f(z_{(m)}), \textrm{  where} \ \ z_{(m)} \sim \textrm{Cat}(w_{(m)}^{1:K}, z_{(m)}^{1:K}),
\end{align*}
we see that these resemble importance sampling estimators, with normalized weights $\frac{\hat{C}_{(m)}}{\sum_{m=1}^M \hat{C}_{(m)}}$, $m=1,\dots, M$. To perform vectorized operations, we thus pick a value $M^*$ and for each observation $x_j$ in a minibatch, we sample (with replacement) $M^*$ samplers from the $M_j$ available, each with probability $\frac{\hat{C}_{(m)}}{\sum_{m=1}^M \hat{C}_{(m)}}$. The resulting $M^*$ SMC samplers are then used to construct each of these two gradient estimators. This results in consistent dimensions for vectorizing minibatch computations of gradients. Of course, if LT-SMC runs are cheap, the procedure can be performed naively as in \Cref{algorithm:smc_wake} where $n$ LT-SMC runs are performed at every gradient step. One additional alternative that also saves computation is to increment $M$ less frequently, e.g., running LT-SMC for all $n$ points but only every 10 gradient steps. 

In practice, the estimates $\hat{C}_{(m)}$ of the normalization constant may have large variance, especially in high-dimensional problems. This can introduce sample deficiency into the SMC-Wake algorithm, as all weight may concentrate on one $\hat{C}_{(m)}$, and thus effectively particles from a single sampler are used. In implementation, this difficulty can be alleviated by considering only a random subset of SMC samplers (and their estimates of the normalizing constants) of size $M'$ at each gradient step, with $M' < M$. This ensures that samples from many different samplers are used to estimate gradients and retains asymptotic guarantees as long as the size of $M'$ is iteratively increased throughout the procedure.

\section{ASYMPTOTIC ANALYSIS}
\label{appendix:consistency}
In this section, we prove asymptotic properties of the gradient estimators $\hat{\nabla}^{(a)}, \hat{\nabla}^{(b)}$, and $\hat{\nabla}^{(c)}$ and also discuss the memory requirements of each. 

Following \citet{Naesseth2019Elements}, we let $\mathbf{u}$ denote all random variables generated by a run of likelihood-tempered SMC (\Cref{algorithm:LikelihoodTemperedSMC}) with a fixed temperature schedule $\tau_1, \dots, \tau_T$. We have the following result from Proposition 7.4.1 of \citet{DelMoral2004FeynmanKacFormulae}, also given in Theorem 2.1 of \citet{Naesseth2019Elements}, whose notation we follow more closely.

\begin{proposition_star}
    For any integrable test function $f$ and any run of $\Cref{algorithm:LikelihoodTemperedSMC}$, we have
    \begin{equation}
    \label{eqn:unbiasedness_smc}
    \mathbb{E}_{\mathbf{u}} \bigg{[} \prod_{s=1}^T \bigg{(} \frac{1}{K} \sum_{k=1}^K \Tilde{w}_s^k \bigg{)} \sum_{k=1}^K w_T^k f(z_T^k) \bigg{]} = C \mathbb{E}_{p(z \mid x)} f(z)
    \end{equation}
    with $C = p(x)$, where $\Tilde{w}_s$ denotes unnormalized weights and $w_s$ denotes normalized weights at any stage $s$ of the SMC sampler.
\end{proposition_star}

The above result The result holds for a general sequence of unnormalized target distributions $\Tilde{\gamma}_1, \dots \Tilde{\gamma}_T$ and any general SMC sampling algorithm, but we have specialized to the likelihood-tempered case in the above where these unnormalized targets are given by $p(z)p(x \mid z)^{\tau_1}, \dots, p(z)p(x \mid z)^{\tau_T}$ for $0 = \tau_1 < \cdots < \tau_T = 1$. The unbiasedness of the estimator $\hat{C}$ (\autoref{eqn:evidence_estimator}) for $C$ results from the above proposition with the special case $f(z) = 1$. We now wish to show the consistency or asymptotic unbiasedness of the estimators below, reproduced from \Cref{tab:estimators}.

\begin{align*}
    \hat{\nabla}^{(a)} &= \frac{\frac{1}{M}\sum_{m=1}^M \hat{C}_{(m)} \big{(} \sum_{k=1}^K w^k_{(m)} f(z^k_{(m)}) \big{)}}{\frac{1}{M}\sum_{m=1}^M \hat{C}_{(m)}} \\
    \hat{\nabla}^{(b)} &=  \frac{\frac{1}{M}\sum_{m=1}^M \hat{C}_{(m)}  f(z_{(m)})}{\frac{1}{M}\sum_{m=1}^M \hat{C}_{(m)}}, \textrm{  where  } z_{(m)} \sim \textrm{Cat}(w_{(m)}^{1:K}, z_{(m)}^{1:K})\\
   \hat{\nabla}^{(c)} &= \frac{\hat{C}_{(M)} \big{(} \sum_{k=1}^K w^k_{(M)} f(z^k_{(M)})}{\frac{1}{M} \sum_{m=1}^M \hat{C}_{(m)}}\\
\end{align*}

We can now consider each of the three estimators in turn, as we prove \Cref{thm:consistency}, reproduced below. 

\begin{customprop}{3}
    For an observation $x_j$, suppose that $f(z) = -\nabla_\phi \log q_\phi(z \mid x_j)$ is a bounded and measurable function for any $\phi \in \Phi$. For $m=1,\dots, M$, let the random variables $\hat{C}_{(m)}$,  $w^{1:K}_{(m)}$, and $z^{1:K}_{(m)}$ result from an independent run of LT-SMC (\Cref{algorithm:LT-SMC_short}) with a fixed temperature schedule. Then, the gradient estimators $\hat{\nabla}_j^{(a)}$ and $\hat{\nabla}_j^{(b)}$ are strongly consistent for $\mathbb{E}_{p(z \mid x_j)} f(z)$ as $M \to \infty$ and the estimator $\hat{\nabla}_j^{(c)}$ is asymptotically unbiased as $M \to \infty$.
\end{customprop}
\begin{proof}
We first note that by the strong law of large numbers, the denominator $\frac{1}{M}\sum_{m=1}^M \hat{C}_{(m)} \overset{a.s.}{\to} C$. This holds immediately because each SMC sampler is generated independently; therefore, the random variables $\mathbf{u}$ generated by each run are independent. In particular, the random variables $\hat{C}_{(m)}$ are independent for all $m$, which implies the result. Below, we discuss each gradient estimator in turn. 

    \textbf{(a)} By \Cref{eqn:unbiasedness_smc} and the definition of $\hat{C}_{(m)}$ (cf. \Cref{eqn:evidence_estimator}), the numerator converges almost surely to $C \mathbb{E}_{p(z \mid x)} f(z)$ by the strong law of large numbers by similar logic to the above. Accordingly, by the continuous mapping theorem the quotient \[\hat{\nabla}^{(a)} = \frac{\frac{1}{M}\sum_{m=1}^M \hat{C}_{(m)} \big{(} \sum_{k=1}^K w^k_{(m)} f(z^k_{(m)}) \big{)}}{\frac{1}{M}\sum_{m=1}^M \hat{C}_{(m)}}\] converges almost surely to $\mathbb{E}_{p(z \mid x)} f(z)$ and is therefore strongly consistent. An important detail that allows application of the continuous mapping theorem is that the random variables $\hat{C}_{(m)}$ are strictly positive, i.e., $\hat{C}_{(m)} > 0$, avoiding division by zero issues.

\textbf{(b)} First formally define the random variables conditional on $\mathbf{u}_{(m)}$ for the $m$th sampler by
    \begin{align}
    \hat{C}_{(m)} \mid \mathbf{u}_{(m)} &= \delta(\hat{C}_{(m)}) \\
    z_{(m)} \mid \mathbf{u}_{(m)} &\sim \textrm{Cat}(w_{(m)}^{1:K}, z_{(m)}^{1:K}).
\end{align}

In other words, conditional on $\mathbf{u}_{(m)}$, the random variable $\hat{C}_{(m)}$ is a point mass at the quantity defined in \autoref{eqn:evidence_estimator} and the random variable $z_{(m)}$ is a draw from the final particle distribution at stage $T$ for the $m$th sampler. Then we have by the law of iterated expectation for any $m$
\begin{align*}
    \mathbb{E} \big{[} \hat{C}_{(m)} f(z_{(m)}) \big{]} &= \mathbb{E}_{\mathbf{u}_{(m)}} \bigg{[} \mathbb{E} \big{[}  \hat{C}_{(m)} f(z_{(m)}) \mid \mathbf{u}_{(m)} \big{]}\bigg{]} \\
    &= \mathbb{E}_{\mathbf{u}_{(m)}}\bigg{[} \hat{C}_{(m)} \mathbb{E} \big{[} f(z_{(m)}) \mid \mathbf{u}_{(m)} \big{]}\bigg{]} \\
    &= \mathbb{E}_{\mathbf{u}_{(m)}}\bigg{[} \hat{C}_{(m)} \sum_{k=1}^K w^k_{(m)} f(z^k_{(m)}) \bigg{]} \\
    &= C \mathbb{E}_{p(z \mid x)} f(z)
\end{align*}
by \Cref{eqn:unbiasedness_smc}, and so again by the strong law of large numbers, the numerator of (b) converges almost surely to $C \mathbb{E}_{p(z \mid x)} f(z)$. By the same reasoning as above, the quotient that defines the ratio estimator $\hat{\nabla}^{(b)}$ converges almost surely to $\mathbb{E}_{p(z \mid x)} f(z)$, and so is strongly consistent for $\mathbb{E}_{p(z \mid x)} f(z)$.

\textbf{(c)} The estimator $\hat{\nabla}^{(c)}$ is asymptotically unbiased, but not consistent because the variance does not tend to zero. Using the fact that $\frac{1}{M}\sum_{m=1}^M \hat{C}_{(m)} \overset{a.s.}{\to} C$ and hence $\frac{1}{M}\sum_{m=1}^M \hat{C}_{(m)} \overset{p}{\to} C$, an application of Slutsky's theorem yields that the distribution of this estimator tends to that of the random variable $\frac{\hat{C}_{(m)}}{C} \sum_{k=1}^K w^k_{(m)} f(z^k_{(m)})$ which clearly has expectation $\mathbb{E}_{p(z \mid x)} f(z)$ by \Cref{eqn:unbiasedness_smc}. This suffices to show asymptotic unbiasedness (cf. \citet{LehmannCasella1998TheoryofPointEstimation} pg. 438, Definition 2.1).

\end{proof}

The gradient estimators $\hat{\nabla}^{(a)}$ and $\hat{\nabla}^{(b)}$ may be viewed as performing a form of ``meta importance sampling'' or ``distributed importance sampling'' \citep{Naesseth2019Elements}, where the ``weights'' are the evidence estimates $\hat{C}_{(m)}$ across $M$ different samplers, properly normalized to sum to unity by the denominator. Keeping in mind that $\hat{C}_{(m)}$ estimates the evidence $p(x)$ for a given observation $x$, higher weights thus correspond to particle sets for which the estimator of the evidence is higher---this provides an intuitive understanding of which SMC samplers receive higher weight. The unnormalized weights are computed based on $p(z, x)$, so higher weights indicate particles that better explain the observed data. 

The numerator of the gradient estimator $\hat{\nabla}^{(c)}$ has variance that does not tend to zero, and therefore this cannot result in a consistent estimator. However, the variance of this estimator can be reduced at the standard Monte Carlo rate by averaging over some finite number of SMC sampler runs $L < M$ using a rolling window approach, e.g., averaging over the most recent 10 SMC sampler runs. 

We conclude with a brief discussion of the memory requirements of the gradient estimators. First, observe that the quantity
\begin{equation*}
    \frac{1}{M} \sum_{m=1}^M \hat{C}_{(m)} = \frac{M-1}{M}\left(\frac{1}{M-1} \sum_{m=1}^{M-1} \hat{C}_{(m)}\right) +  \frac{1}{M} \left( \hat{C}_{(M)} \right)
\end{equation*}
and so this average can be maintained with only $O(1)$ memory as $M \to \infty$. Considering the three estimators in turn, for each data point $x$ we thus immediately have that $\hat{\nabla}^{(a)}$ requires $O(MK)$ memory by maintaining a $K$-particle set for each of the $M$ sampler runs. By resampling a single particle and applying the law of iterated expectation, the estimator $\hat{\nabla}^{(b)}$ just requires $O(M)$ memory for $M$ sampler runs.  The estimator $\hat{\nabla}^{(c)}$ is the cheapest---it only uses the $K$-particle set from the most recent sampler, i.e. the $M$th, to compute the numerator, and so thus requires $O(K)$ memory. Of course, to repeatedly construct the samplers for increasing $M$, the SMC-Wake procedure also requires sufficient memory to run LT-SMC itself which is $O(K)$ for every run.

\section{TWO MOONS}
\label{appendix:two_moons}

We generate 100 points $x_1, \dots, x_{100}$ independently from the generative model in \Cref{experiments:two_moons}. The encoder architecture is a neural spline flow (NSF) \citep{Durkan2019NeuralSplineFlow}, adapted from the \texttt{sbi} Python library \citep{TejeroCantero2020SBIPython}. We use the default NSF settings as the problem is low-dimensional enough to avoid embedding nets, etc. For all experiments, some tensor operations in PyTorch were performed using NVIDIA GeForce RTX 2080 Ti graphical processing units (GPUs). 

All methods were trained using a batch size of 16, with a learning rate of 0.0001 for 50,000 gradient steps. We use a large set of $K=1000$ particles for all the methods. SMC-Wake reruns one SMC sampler (corresponding to one observation) at random every 10 gradient steps, corresponding to about $M=50$ runs of the SMC sampler per point throughout the fitting procedure. Even relatively small values for $M$ such as those used here yield a significant improvement over wake-phase training. For all SMC runs, the proposal distribution was a 5-step Metropolis-Hastings random-walk kernel with Gaussian steps and $\sigma^2=0.1^2$ as its variance. For wake-phase updates, instances where the self-normalized importance sampling estimate of the gradient was undefined were ignored and re-tried.

\begin{figure}[ht!]
  \centering
  \includegraphics[width=.7\linewidth]{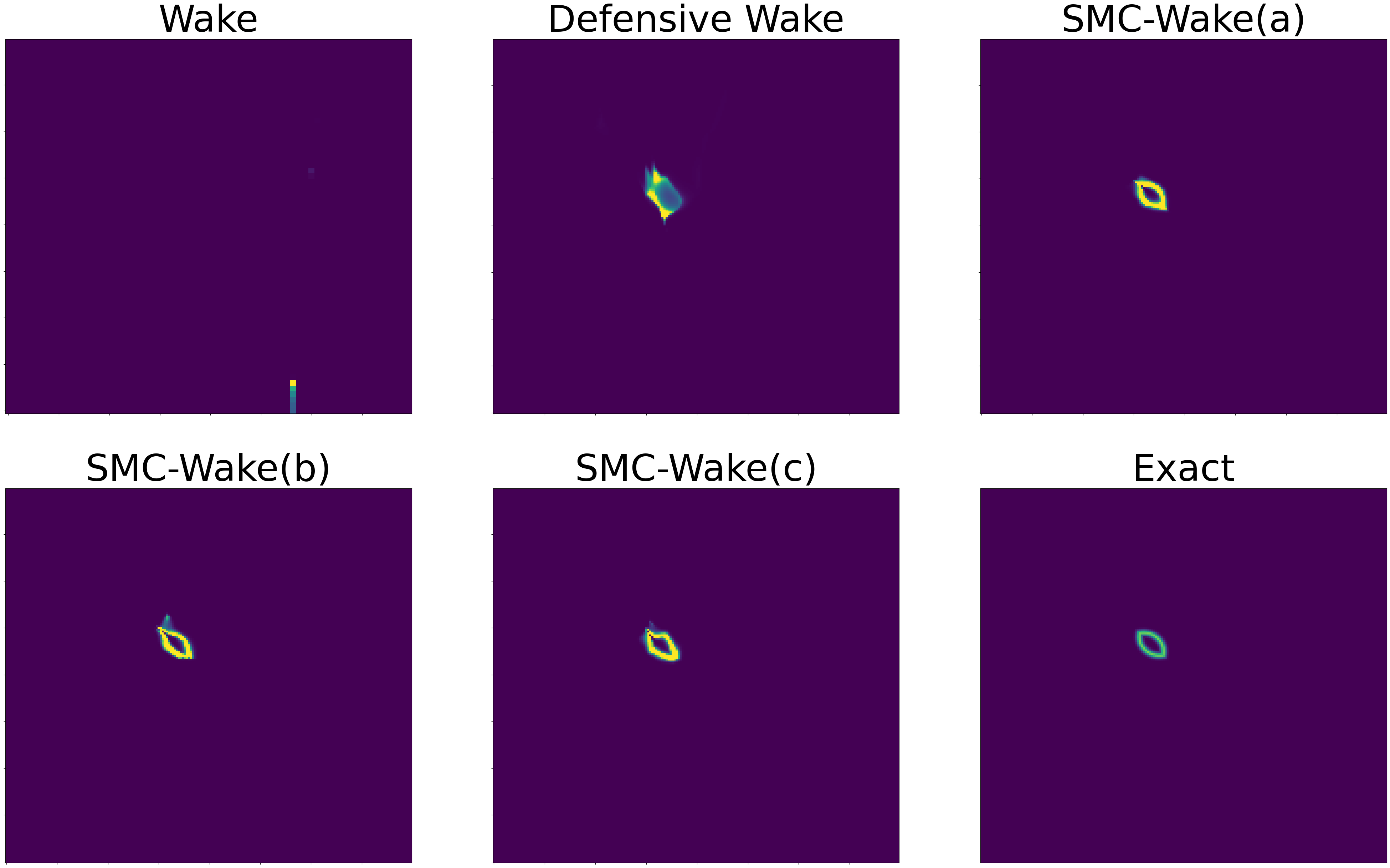}
  \caption{Comparison of two-moons variational posteriors.}
  \label{fig:two_moons_comp_appendix1}
\end{figure}

\begin{figure}[ht!]
  \centering
  \includegraphics[width=.7\linewidth]{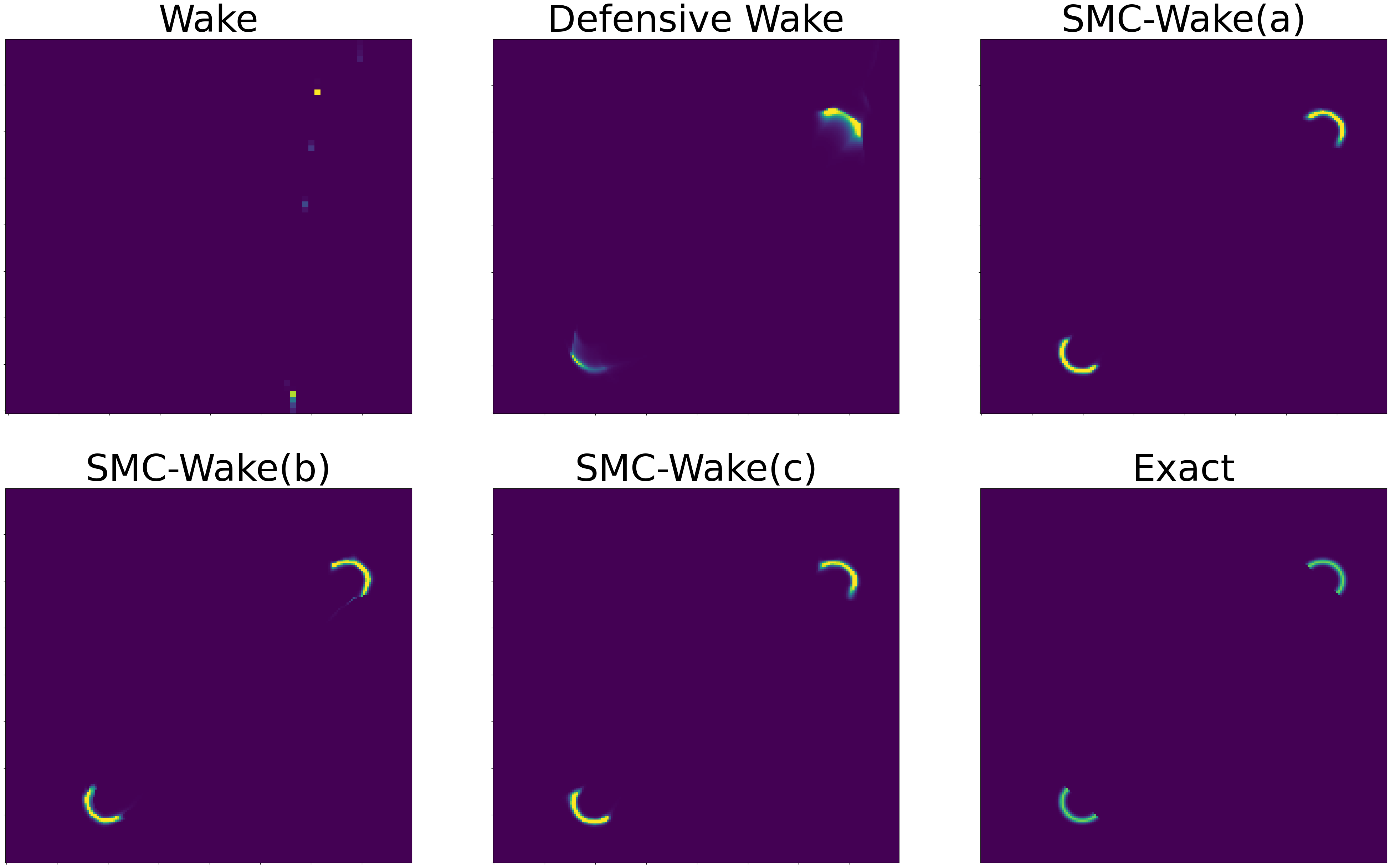}
  \caption{Comparison of two-moons variational posteriors.}
  \label{fig:two_moons_comp_appendix2}
\end{figure}

\begin{figure}[ht!]
  \centering
  \includegraphics[width=.7\linewidth]{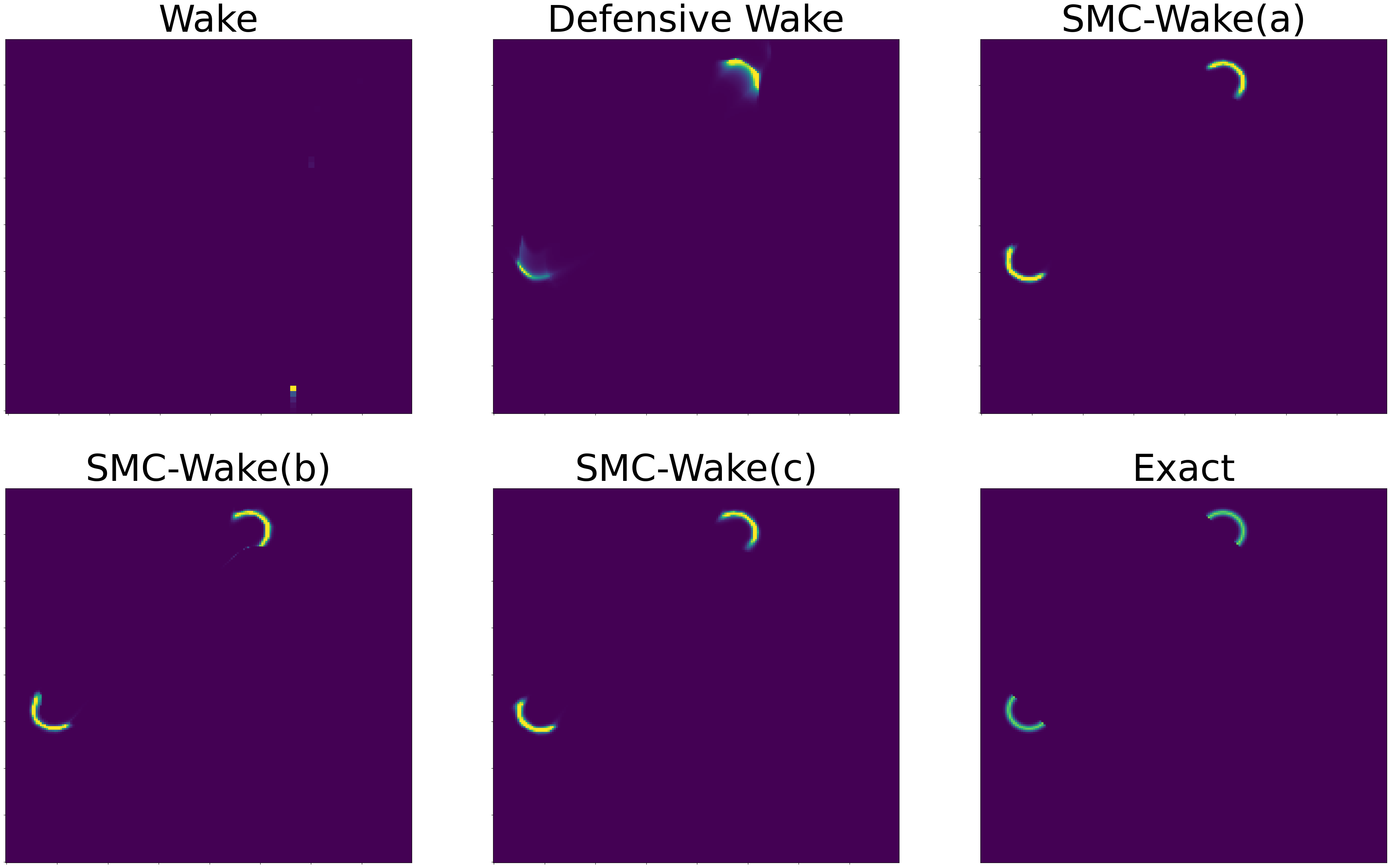}
  \caption{Comparison of two-moons variational posteriors.}
  \label{fig:two_moons_comp_appendix3}
\end{figure}

In the same manner as in \Cref{fig:moon_plots}, we plot the variational approximations for several other points in the training set in \Cref{fig:two_moons_comp_appendix1}, \Cref{fig:two_moons_comp_appendix2}, and \Cref{fig:two_moons_comp_appendix3}. We show results based on training via the three gradient estimators $\hat{\nabla}^{(a)}, \hat{\nabla}^{(b)}$, and $\hat{\nabla}^{(c)}$, denoted as SMC-Wake(a)-(c), respectively. 

\begin{figure}[ht!]
  \centering
  \includegraphics[height=250pt, width=.7\linewidth]{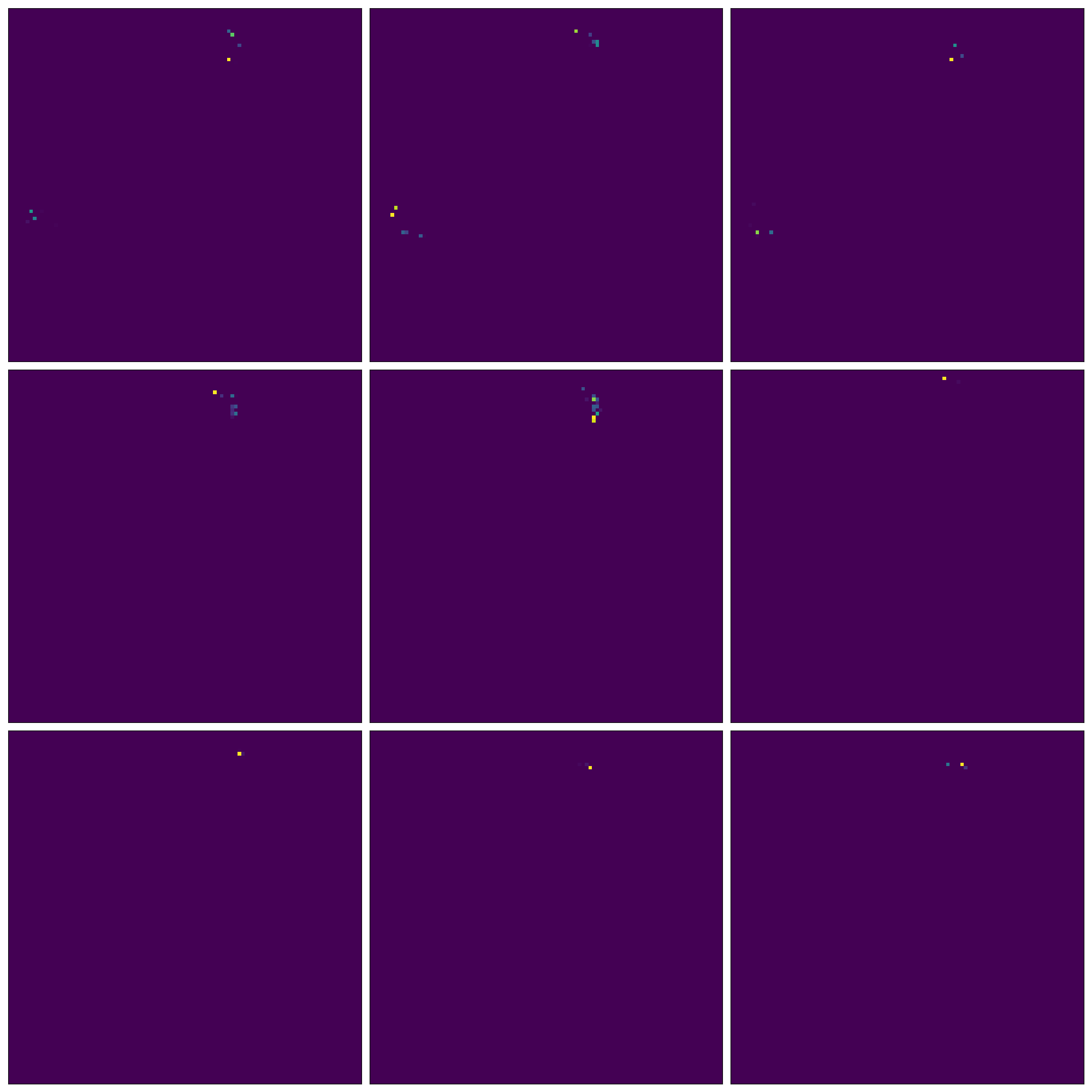}
  \caption{Visualizations of the importance sampling particle approximations with $K=1000$ for wake-phase training. The nine panels are sampled from the inference network at every hundred gradient steps, so at step $100, 200,\dots, 900$. We see that even within the first one thousand gradient steps, mass is concentrated.}
  \label{fig:wake_mass_concentrate_two_moons}
\end{figure}

Clearly, wake-phase training concentrates mass. We show that this concentration of mass occurs very rapidly, in this case within one thousand gradient steps, in \Cref{fig:wake_mass_concentrate_two_moons}.

\section{MNIST}
\label{appendix:mnist}

Amortized inference is often used to simultaneously train a model (decoder) and a posterior approximator (encoder).  In this setting, latent representations can be considered as compressed versions of data, from which reconstructions should nevertheless be accurate \citep{Kingma2019VAEIntro,Burda2016IWAE}. Given 1000 MNIST digits $x$ and labels $\ell$ denoted by $\{x_i, \ell_i\}_{i=1}^{1000}$, we simultaneously fit an encoder $q_\phi(z \mid x, \ell)$ and a decoder $p_\psi(x \mid \ell, z)$.

Each class $\ell$ is modeled by a sigmoidal belief network (SBN) \citep{Saul1996SigmoidBeliefNetwork}, and thus the model is given by
\begin{equation}
    p_\psi(x \mid \ell, z) \sim \mathcal{N}(\sigma\big{(}W_\ell z + b_\ell\big{)}, \tau^2 I_d)
\end{equation}
where $\sigma(\cdot)$ denotes the sigmoid function. The observations $x$ are normalized to have intensities in $[0,1]$, and therefore the sigmoidal transformation describes the data well. Setting $\tau=0.01$ results in a highly peaked likelihood.

We take $p=16$ as the latent dimension, and thus $W_\ell \in \mathbb{R}^{p \times 784}$ and $b_\ell \in \mathbb{R}^{784}$ for all $\ell$. For the inference network, we use a simple amortized diagonal Gaussian distribution $q_\phi(z \mid x) = \mathcal{N}(\mu(x), \textrm{diag}(\sigma(x)))$  whose parameters are the output of a three-layer dense network with ReLU activation and hidden dimension of 64. We use logarithmic scales for numerical stability. The prior $p(z)$ is given by a standard multivariate Gaussian distribution. We used a learning rate of 0.0005 to fit both the decoder and encoder. All methods alternated an update to the model parameters followed by an update to the encoder parameters, with 50,000 steps each. 

\begin{figure}[ht!]
\centering
\includegraphics[width=.7\linewidth]{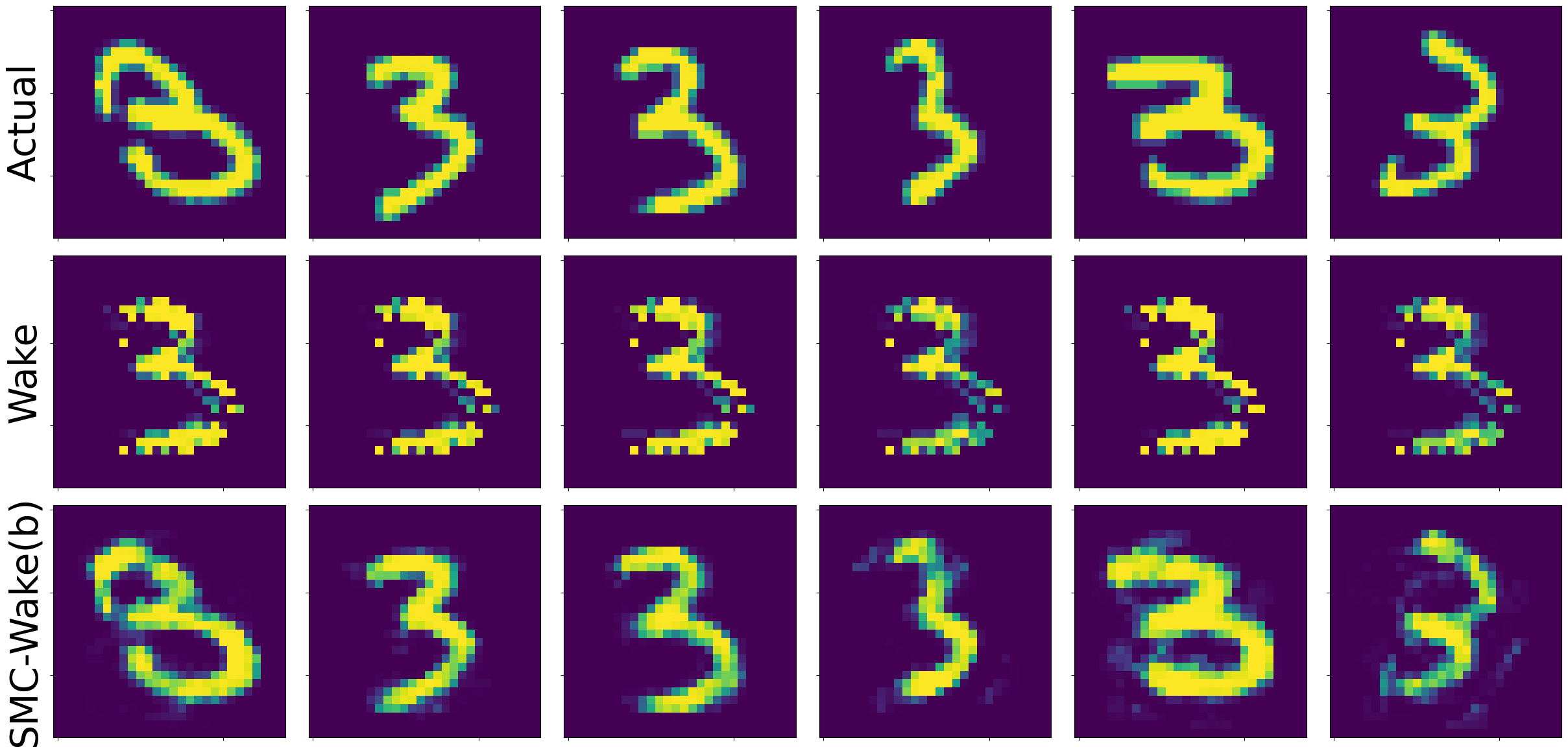}
\caption{}
\label{mnist_digit_3}
\end{figure}

\begin{figure}[ht!]
\centering
\includegraphics[width=.7\linewidth]{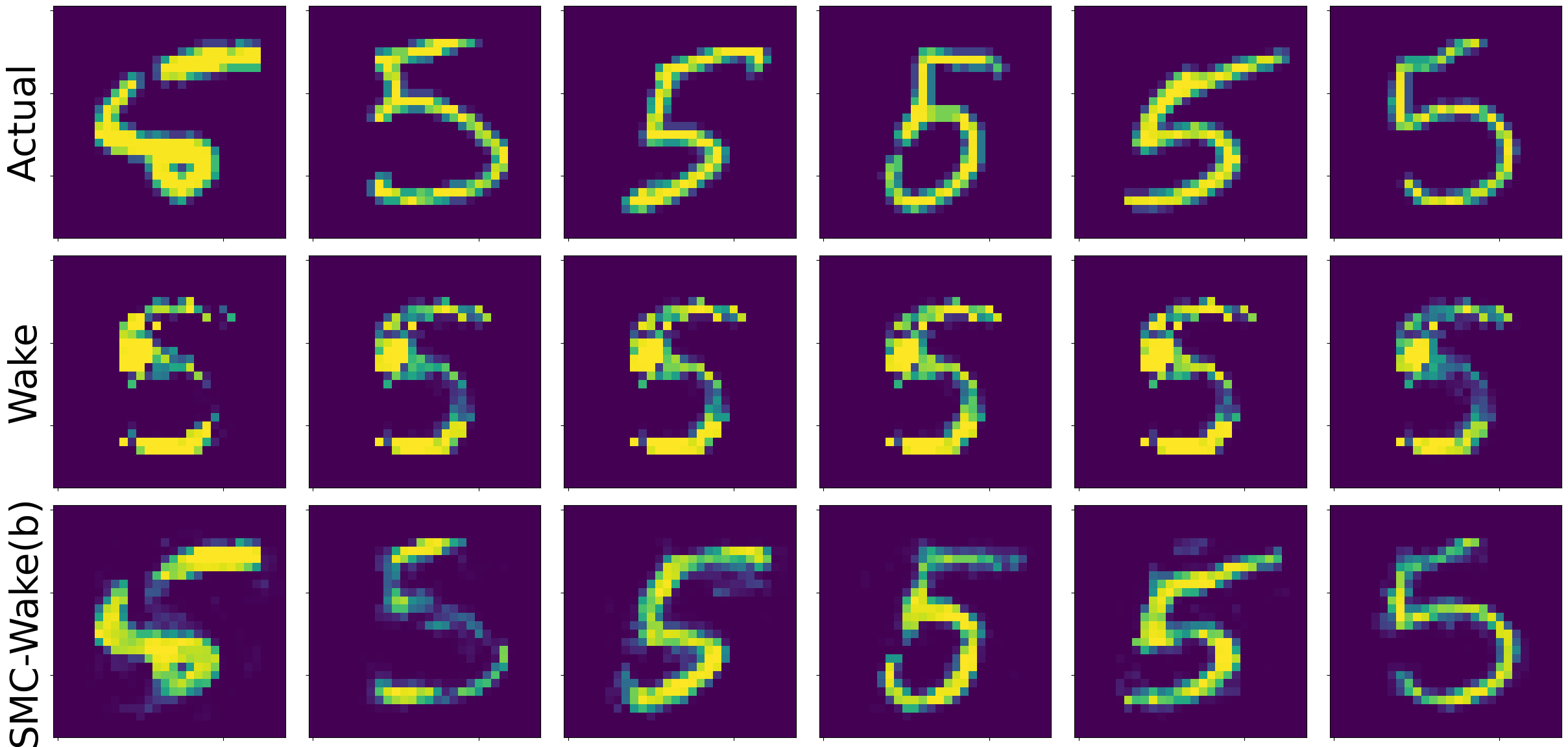}
\caption{}
\label{mnist_digit_5}
\end{figure}

\begin{figure}[ht!]
\centering
\includegraphics[width=.7\linewidth]{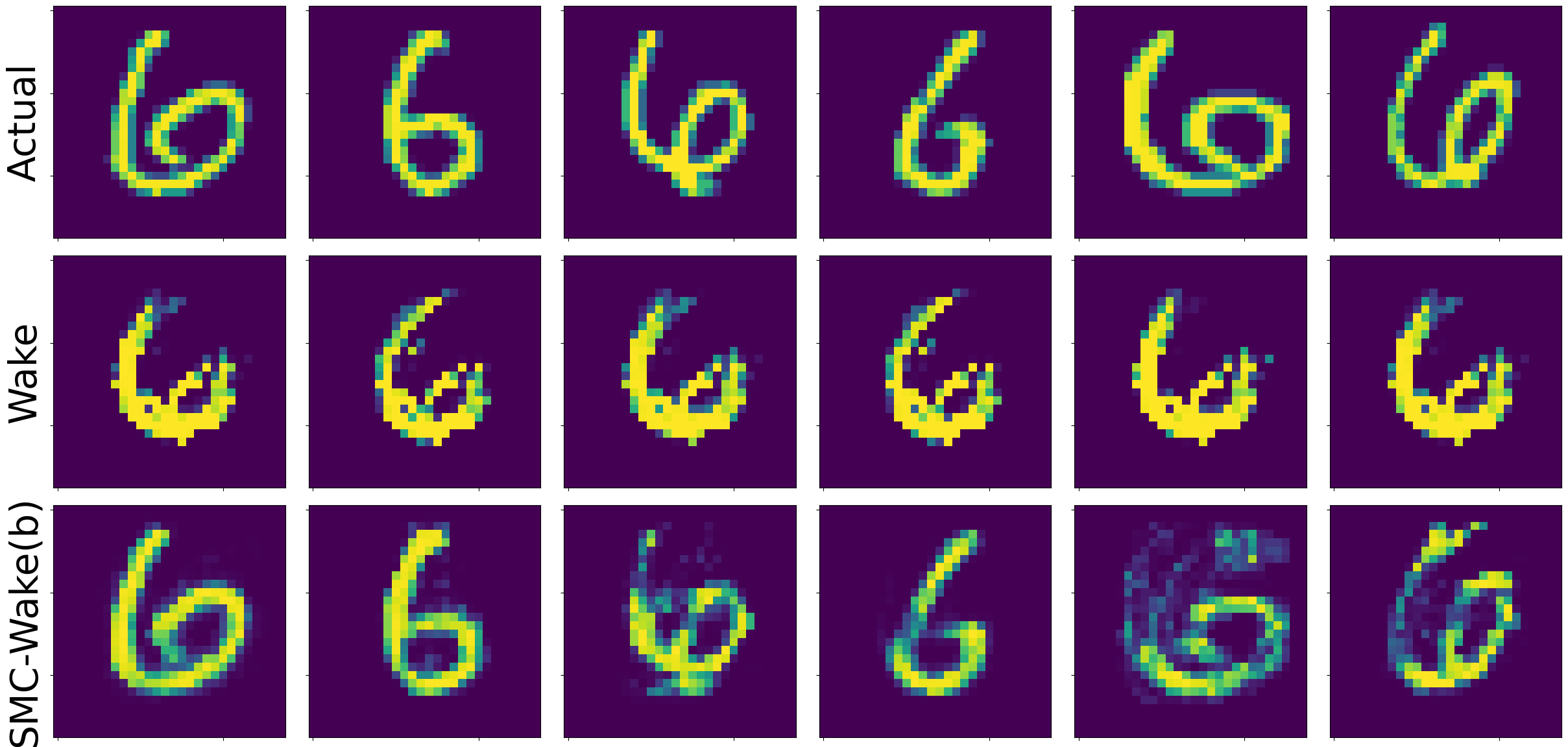}
\caption{}
\label{mnist_digit_6}
\end{figure}

The model parameters $\phi$ are fit to maximize the IWBO while the encoder is fit by wake-phase training and SMC-Wake, using the gradient estimator $\hat{\nabla}^{(b)}$. In addition to \Cref{fig:mnist_reconstructions}, below we plot several similar plots for other digits in \Cref{mnist_digit_3}, \Cref{mnist_digit_5}, and \Cref{mnist_digit_6}. 

The procedure is implemented by alternating gradient updates to $\theta$ and to $\phi$, respectively. This can be done safely without losing the asymptotic guarantees of the estimators $\hat{\nabla}^{(a)},\hat{\nabla}^{(b)}$, and $\hat{\nabla}^{(c)}$ provided that the model parameters $\theta$ converge (stop changing) to some $\theta^*$ by some finite $M^* < \infty$. In this case, as $M \to \infty$ the average of normalization constant estimates still tends to the normalization constant for the correct model (i.e. with parameters $\theta^*$). This follows directly from Kolmogorov's generalized LLN for independent (but not identically distributed) r.v.'s: $(\frac{1}{M}\sum_{m=1}^M \hat{C}_{(m)}) - (\frac{1}{M}\sum_{m=1}^M \mu_{(m)}) \overset{a.s.}{\to} 0$, where $\mu_{(m)} = \mathbb{E} \hat{C}_{(m)}$. If $\theta = \theta^*$ for all $M > M^*$,  then $\frac{1}{M}\sum_{m=1}^M \mu_{(m)} \to C$, the normalization constant for the final model with parameters $\theta^*$. This is true even without discarding early normalization constant estimates, although discarding may be desirable in practice in some applications. Intuitively, the effect of finitely many normalization constant estimates for ``wrong'' models diminishes as we aggregate infinitely many estimates for the ``correct'' model.

\section{GAUSSIAN LINEAR MODEL}
\label{appendix:gaussian}

For the model
\begin{align*}
  z &\sim \mathcal{N}(0, \sigma^2 I_p) \\
  x &\mid z \sim \mathcal{N}(Az, \tau^2 I_d)
\end{align*}
the posterior distribution is proportional to
\begin{align*}
  &\exp \bigg{(}-\frac{1}{2}\bigg{[} \frac{z^\top z}{\sigma^2} + \frac{(x-Az)^\top(x-Az)}{\tau^2} \bigg{]}\bigg{)} \\
  \propto \ \ &\exp \bigg{(}-\frac{1}{2}\bigg{[} z^\top \big{(} \frac{I_d}{\sigma^2} + \frac{A^\top A}{\tau^2 }\big{)}z -2\frac{x^\top Az}{\tau^2} \bigg{]}\bigg{)}\\
  = & \exp \bigg{(}-\frac{1}{2}\bigg{[} z^\top M z -2b^\top z \bigg{]}\bigg{)}\\
  = & \exp \bigg{(}-\frac{1}{2}\bigg{[} \left(z-M^{-1} b\right)^T M\left(z-M^{-1} b\right)-b^T M^{-1} b \bigg{]}\bigg{)}\\
  \propto \ \ & \exp \bigg{(}-\frac{1}{2}\bigg{[} \left(z-M^{-1} b\right)^T M\left(z-M^{-1} b\right) \bigg{]}\bigg{)}.
\end{align*}
by a completing the square argument. Therefore, the posterior distribution is $\mathcal{N}(M^{-1}b, M^{-1})$ where
\begin{align*}
  M = \frac{I_d}{\sigma^2} + \frac{A^\top A}{\tau^2 } \in \mathbb{R}^{d \times d}
\end{align*}
and
\begin{align*}
  b = \frac{A^\top x}{\tau^2} \in \mathbb{R}^d.
\end{align*}

The experimental details for the nested MCMC examples are as follows: as stated,  we use dimensions $p=50$ and $d=100$, as well as $\sigma = \tau = 1$, and the design matrix $A \in \mathbb{R}^{d \times p}$ is fixed. The encoder parameterizes a multivariate Gaussian distribution in $p$ dimensions. The variational family $q_\phi(z \mid x) \sim \mathcal{N}(\mu, L L^\top + \epsilon I)$ is flexible enough to approximate the exact posterior. We add the additional $\epsilon I$ term with $\epsilon = 0.0001$ for numerical stability, but this does not overly constrain the variational family in this case, as the true posterior covariance has eigenvalues at least this large. The variational parameters given any observation are the outputs of a dense neural network with 4 dense layers each of dimension 64, with ReLU activation. For each run of LT-SMC, each stage $t$ performs the mutation step using 100 steps of a Metropolis-Hastings random-walk kernel with $\sigma = 0.01$, each kernel leaving the stage $t$ target $\gamma_{t}$ invariant.  

We fit the encoder using the SMC-PIMH-Wake objective using a batch size of 32 data points per gradient step (out of $n=50$ possible in the given dataset), with learning rate 0.0001 for 40,000 gradient steps. The number of particles used is $K=100$. After each gradient step, we randomly pick just one $x_j \in \mathcal{D}$, rerun LT-SMC for it, and take a Metropolis-Hastings step within the outer PIMH loop for that particular $x_j$. SMC samplers are more expensive to run than SNIS, and we found that rerunning the SMC samplers for each point in $\mathcal{D}$ at each gradient step was unnecessary to achieve good results. Across 40,000 gradient steps, this averages to about 800 MCMC steps for each $x_j \in \mathcal{D}$.

We compare to Markovian Score Climbing (MSC), using $q_\phi$ as the proposal distribution within a CIS kernel for MCMC. As IS is more lightweight than SMC, we perform an MCMC step for each of the $n=50$ Markov chains (one for each datapoint) every time its corresponding observation $x_j$ is in a minibatch for a gradient step. As the batch size is 32 in our implementation, this corresponds to over 250,000 MCMC steps for each of the 50 chains over a total of 500,000 gradient steps. All other hyperparameters for the fitting procedure are the same as those for SMC-PIMH-Wake, including the number of particles $K$ and the learning rate.

\begin{figure}
    \centering
    \includegraphics[width=.7\linewidth]{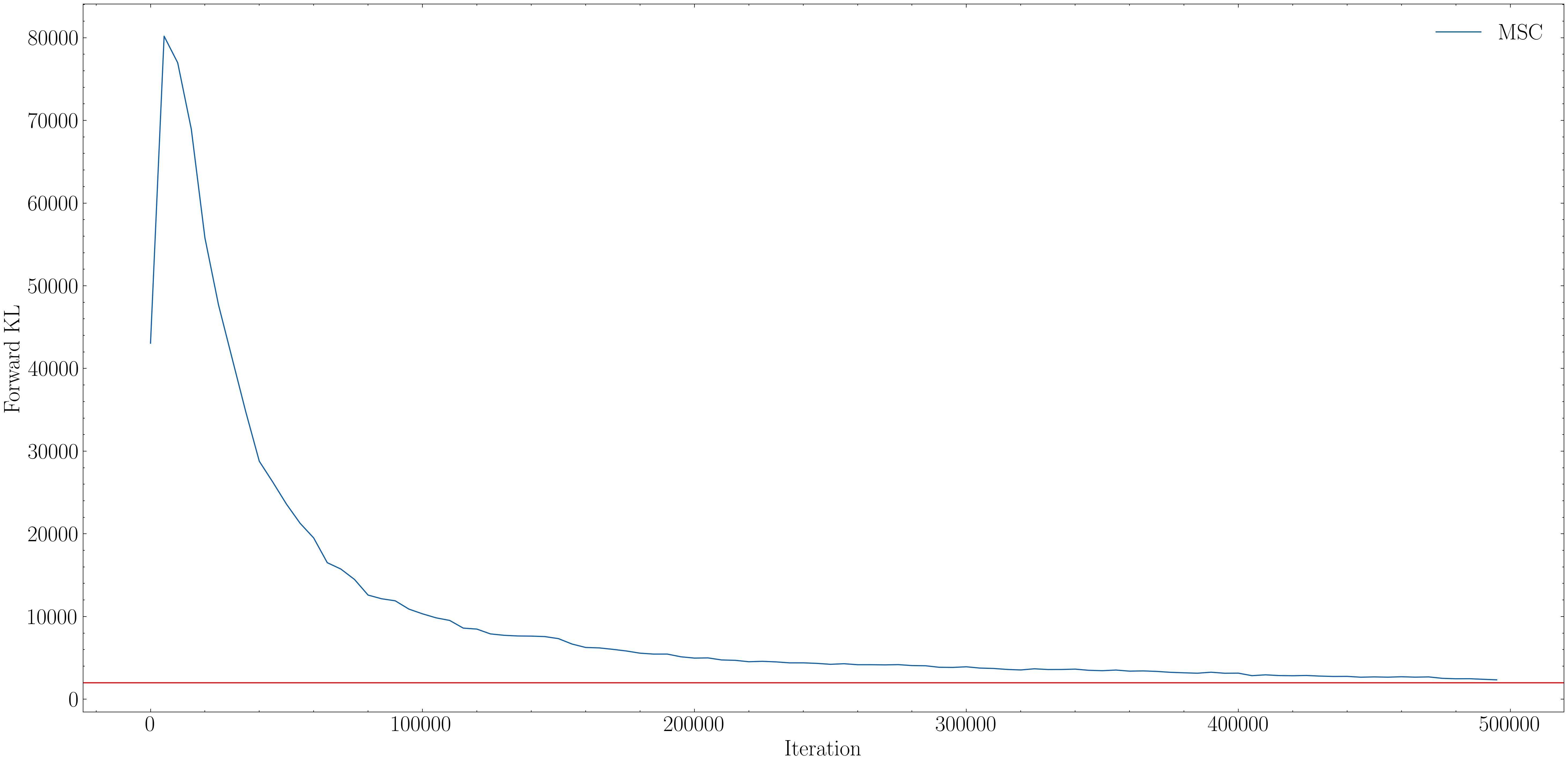}
    \caption{The forward KL divergence averaged across all $n$ observations, plotted throughout the 500,000 gradient steps of fitting by Markovian score climbing (MSC). At convergence, the minimal average forward KL achieved seems to be approximately 5000 or so.}
    \label{fig:msc:convergence}
\end{figure}

In addition to the results of \Cref{tab:gaussian_gaussian_results}, in \Cref{fig:msc:convergence} we plot the average forward KL divergence (recorded every 5,000 gradient steps) throughout the MSC fitting procedure. These results suggest that the fitting procedure for $q_\phi$ is near convergence after 500,000 steps and has converged to an encoder network with an average forward KL of around 2,000 (the red horizontal line in \Cref{fig:msc:convergence} is at 2,000). SMC-PIMH-Wake, on the other hand, fits an encoder with a significantly lower forward KL. 

For the example comparing many samplers and a single large sampler, we have the same setup and architectures as above, except we only use 10 data points for ease, as computing the $K=10,000$ particle sampler for each is computationally expensive. To make the problem more challenging, we also use a different fixed matrix $B$ instead of $A$ that results in a more poorly conditioned posterior, i.e. one that cannot be described exactly by the variational family, and we use a Metropolis-Hastings random-walk kernel with only 10 steps and with a larger $\sigma$ (0.1) to ensure adequate exploration of the space. For each point, a single SMC sampler is run with $K=10,000$ particles, along with $100$ different runs with $K=100$ particles. All runs use a 10-step MHRW transition kernel at each stage similar to that described above. 

We fit the encoder network using a naive form of the SMC-Wake gradient estimator for this setting, as the $K=10,000$ case only has $M=1$ sampler to work with. For the case of $K=10,000$ particles, the gradient is thus estimated at every step using the same sets of particles $z^{1:K} \in \mathbb{R}^{10,000}$ and weights $w^{1:K} \in \mathbb{R}^{10,000}$, as the sampler is run only once. For the $K=100$ case, at each gradient step we select one of the $M=100$ samplers at random, and use the vectors $z^{1:K}, w^{1:K} \in \mathbb{R}^{100}$ to estimate the gradient. This can be thought of as the most naive form of SMC-Wake, as the samplers have even weight and evidence estimates $\hat{C}^{(m)}$ are not used to weight the samplers. Alternatively, this can be considered as the lightweight form of SMC-Wake with estimator $\hat{\nabla}^{(a)}$ focused on sample replenishment with trivial $M' = 1$ (see \Cref{appendix:algo}). As \Cref{fig:many_samplers_vs_one} illustrates, the $K=10,000$ sampler is inefficient: because the LT-SMC procedure is difficult to tune in this large dimension, this sampler has $\mathrm{ESS} \approx 1$, so the larger particle budget does not help much compared to a larger number of SMC samplers.  

\section{GALAXY SED EMULATOR}
\label{appendix:sed}

The Probabilistic Value-Added Bright Galaxy Survey simulates synthetic spectra for the Dark Energy Spectroscopic Instrument (DESI) Bright Galaxy Survey. Its code is freely available at \url{https://github.com/changhoonhahn/provabgs} \citep{Hahn2022Provabgs, Abareshi2022DESIOverview}, and used with permission of the MIT License.

The simulator uses 12 parameters to produce synthetic spectra. We describe these below, reproduced from \citet{Hahn2022Provabgs}.

\resizebox{\linewidth}{!}{$\displaystyle
\begin{array}{|c|c|c|}
\hline \text {Name} & \text {Description} & \text {Prior} \\
\hline \log M_* & \text { log galaxy stellar mass } & \text { uniform over }[7,12.5] \\
\hline \beta_1, \beta_2, \beta_3, \beta_4 & \text { NMF basis coefficients for SFH } & \text { Dirichlet prior } \\
\hline \text { fburst } & \text { fraction of total stellar mass formed in starburst event } & \text { uniform over }[0,1] \\
\hline t_{\text {burst }} & \text { time of starburst event } & \text { uniform over }[10 \mathrm{Myr}, 13.2 \mathrm{Gyr}] \\
\hline \gamma_1, \gamma_2 & \text { NMF basis coefficients for } \mathrm{ZH} & \log \text { uniform over }\left[4.5 \times 10^{-5}, 1.5 \times 10^{-2}\right] \\
\hline \tau_{\mathrm{BC}} & \text { Birth cloud optical depth } & \text { uniform over }[0,3] \\
\hline \tau_{\mathrm{ISM}} & \text { diffuse-dust optical depth } & \text { uniform over }[0,3] \\
\hline n_{\text {dust }} & \text { Calretti }(2001) \text { dust index } & \text { uniform over }[-2,1] \\
\hline
\end{array}$
    }

\begin{figure}[ht!]
  \centering
  \includegraphics[width=.5\linewidth]{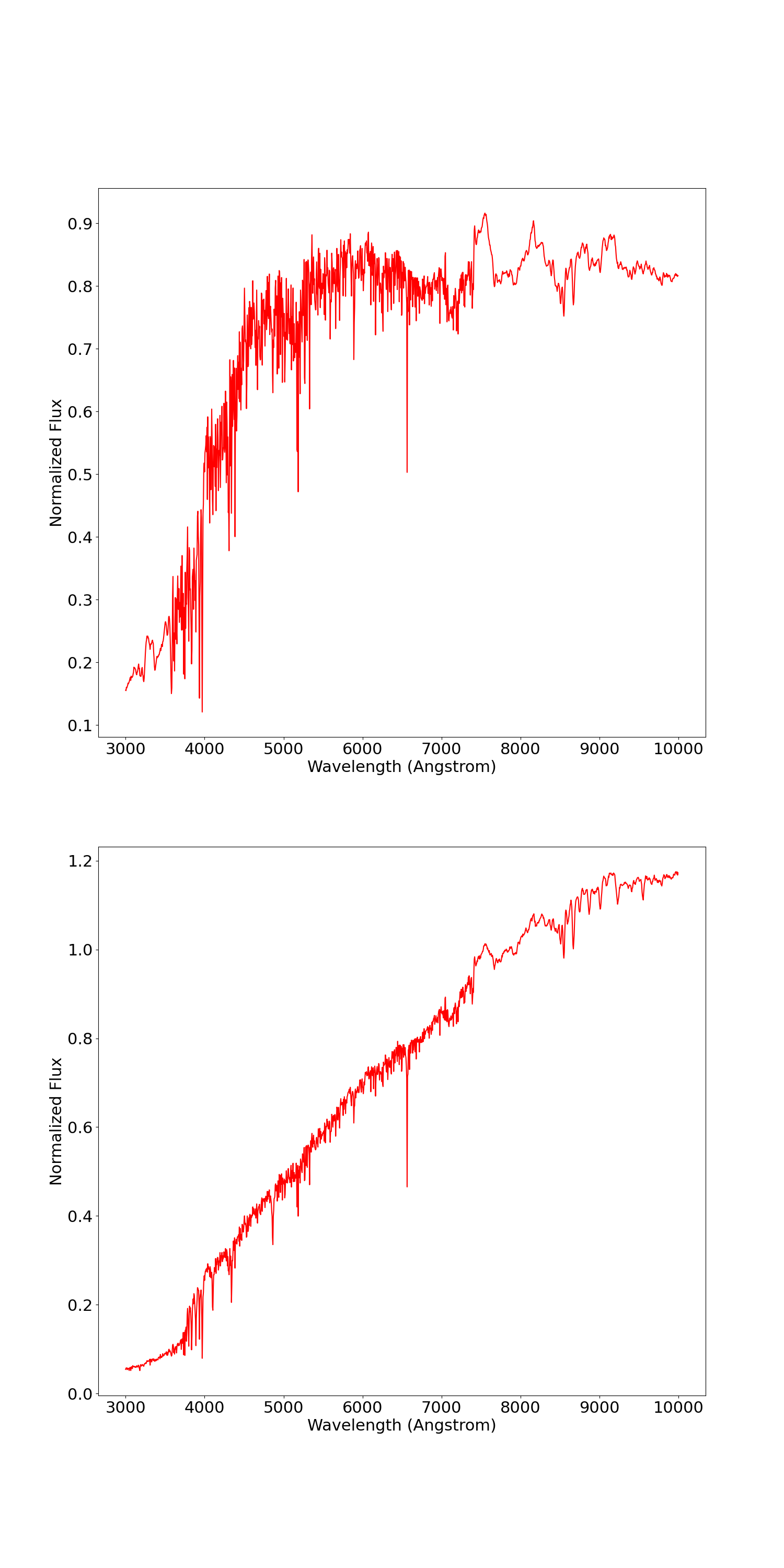}
  \caption{Two examples of draws from our neural network emulator of PROVABGS.}
  \label{fig:example_emulator_draws}
\end{figure}

The simulator is computationally expensive, even using the authors' emulated version; we train our own lightweight neural network emulator of PROVABGS. Because the spectra simulated by PROVABGS are of vastly different magnitudes, we work on the modified problem of normalized spectra, which can be emulated more easily. To train the emulator, we generate batches of synthetic spectra using the prescribed priors, but with fixed $\log M_* = 10.5$, and normalize each to integrate to one. We also resample onto a grid of wavelengths between 3000 and 10,000 Angstrom, with width 5 Angstrom. The emulator is trained with 2000 batches, each consisting of 1000 simulated $(\theta, x)$ pairs. The criterion optimized is the mean-squared error (MSE) with learning rate 0.001. After training, the emulator produces draws (\Cref{fig:example_emulator_draws}) that look similar to the (normalized) PROVABGS outputs. Because we fix the stellar mass parameter, our emulator uses 11 input parameters. 

The uniform and Dirichlet priors present problems for our likelihood-tempered SMC procedure (\Cref{algorithm:LikelihoodTemperedSMC}). As we use Metropolis-Hastings random-walk transition kernels, the boundaries of the support complicate using symmetric Gaussian proposals; random-walks on the simplex are similarly challenging. To resolve this issue, we operate on a transformed space of unconstrained random variables $\tilde{\theta}$ that are bijective transformations of the random variables $\theta$ of interest. Bijections from any uniform range $(a,b)$ to $\mathbb{R}$ can be constructed using scaled logit transformations, and the inverses computed by scaled sigmoidal transforms. Care must be taken with the Dirichlet parameters; these 4 random variables are first transformed into a 3-dimensional space using the warped manifold transformation of \citep{Betancourt2012Simplex}. The resulting random variables reside in $(0,1)$; we can apply a subsequent logit transform from here as above. We perform SMC-Wake to learn distributions on the unconstrained space before applying the transformations to the constrained space for evaluation in \Cref{fig:smcwake_mcmc}. Overall, SMC thus samples on a 10-dimensional unconstrained space before transforming back into an 11-dimensional unconstrained space.

The data in this example are very high-dimensional; even with resampling onto a coarser wavelength grid, spectra are still 1400-dimensional. Adaptive tempering provides one way of efficiently dealing with the likelihood function of such data in a principled way. Most data choose a low initial temperature (on the order of $10^{-5}$ or so) to slowly introduce the likelihood function and prevent overpowering of the prior; manual selection of suitable temperatures in this case may prove more difficult.

We generate 100 observations from our emulator and priors described above and add noise proportional to the signal in each wavelength bin. A neural spline flow (NSF) is used as the amortized variational posterior \citep{Durkan2019NeuralSplineFlow} on $\Tilde{\theta}$. For all methods, we perform 25,000 gradient steps with learning rate 0.0001 using a mini-batch size of 32.

\begin{figure}
    \centering
    \includegraphics[width=.63\linewidth]{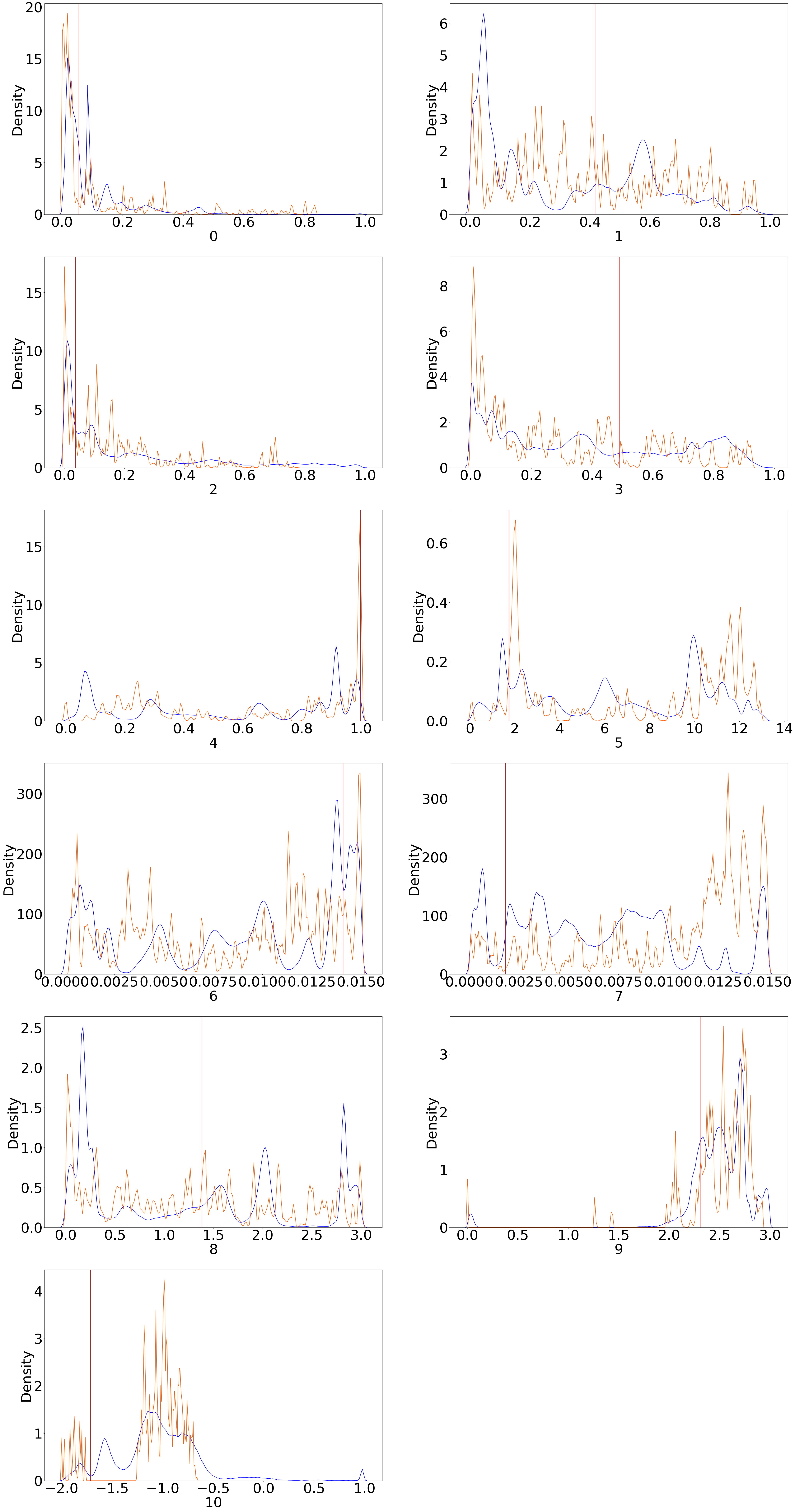}
    \caption{Comparison of SMC-Wake posteriors (blue) to those obtained by MCMC (orange) for a single observation.}
    \label{fig:smc_mcmc_full_11}
\end{figure}

We fit the encoder using SMC-Wake with gradient estimator $\hat{\nabla}^{(a)}$, taking $K=100$ for all LT-SMC runs. The mutation kernel in LT-SMC is a 50-step Metropolis-Hastings kernel with noise parameter $\sigma^2 = 0.1^2$. SMC-Wake performs well with a minimal number of LT-SMC runs---we only rerun LT-SMC for a single point (chosen at random) every $50$ gradient steps. Across the 25,000 gradient steps, this averages to only about $M=5$ LT-SMC runs for each of the $n=100$ observations. In \Cref{fig:smcwake_mcmc} we only showed posteriors for 4 out of 11 parameters for one of the one hundred observations due to space constraints; here we show all 11 in \Cref{fig:smc_mcmc_full_11} for this same point. We also performed training by wake-phase updates for this example, but wake-phase training exhibits a troubling numerical stability issue that impairs training, whereby some parameter values sampled from $q_\phi(\theta \mid x)$ produce spectra with NaN values when fed through the emulator. If this is the case for all $K$ particles sampled from $q_\phi$ in importance sampling, then the wake-phase gradient is undefined. SMC-Wake does not suffer from this issue due to its use of LT-SMC. 

\end{document}